\documentclass[letterpaper, 10pt, journal]{IEEEtran}

\usepackage{cite}
\usepackage{amsmath,amssymb,amsfonts}
\usepackage{mathtools}
\usepackage{amsthm}
\usepackage{graphicx}
\usepackage{textcomp}
\usepackage{xcolor}
\usepackage{bm}
\usepackage{booktabs}
\usepackage{array}
\usepackage{url}
\usepackage{tikz}
\usetikzlibrary{arrows.meta,positioning}
\usepackage{xurl}

\usepackage{stfloats}
\usepackage{enumitem}
\newlist{penumerate}{enumerate}{1}
\setlist[penumerate,1]{label=(P\arabic*),leftmargin=*,itemsep=2pt}
\newlist{aenumerate}{enumerate}{1}
\setlist[aenumerate,1]{label=(\alph*),leftmargin=*,itemsep=2pt}
\newlist{denumerate}{enumerate}{1}
\setlist[denumerate,1]{label=(\alph*),leftmargin=*,itemsep=2pt,start=4}
\theoremstyle{plain}
\newtheorem{assumption}{Assumption}
\newtheorem{theorem}{Theorem}
\newtheorem{lemma}{Lemma}
\newtheorem{proposition}{Proposition}
\newtheorem{corollary}{Corollary}
\theoremstyle{definition}
\newtheorem{definition}{Definition}
\newtheorem{problem}{Problem}
\theoremstyle{remark}
\newtheorem{remark}{Remark}
\usepackage{subcaption}

\newcommand{\norm}[1]{\left\|#1\right\|}
\newcommand{\abs}[1]{\left|#1\right|}
\newcommand{\hpi}{h_{\perp,i}}

\begin{document}
\allowdisplaybreaks
\sloppy

\title{Safe Circumnavigation on Heterogeneous Orbits: A Range-Only HOCBF-QP Framework for Nonholonomic Multi-Agent Systems}

\author{Shubham~Singh, Anand~Singh, Anjali~Tiwari,
Lara~Bri\~n\'on-Arranz, and Katsuki~Fujisawa%
\thanks{S.~Singh, A.~Singh, and K.~Fujisawa are with the Department
of Mathematical and Computing Science, School of Computing,
Institute of Science Tokyo, Tokyo~152-8550, Japan (e-mail:
\texttt{\{shubham.s.ee99, singh.a.ca69, fujisawa.k.2110\}@m.isct.ac.jp}).}%
\thanks{L.~Bri\~n\'on-Arranz is with Univ. Grenoble Alpes, CNRS,
Grenoble INP, GIPSA-lab, 38000 Grenoble, France
(e-mail: \texttt{lara.brinon-arranz@gipsa-lab.grenoble-inp.fr}).}%
\thanks{A.~Tiwari is with the Department of Electrical Engineering,
Indian Institute of Technology Jodhpur, Jodhpur~342030, India
(e-mail: \texttt{anjali.1@iitj.ac.in}).}%
}

\maketitle

\begin{abstract}
Circumnavigating a stationary target with a team of robots
requires each robot to hold its own standoff distance, to stay evenly spaced from its neighbours, and to avoid collisions with the target and with each other at every instant. The task becomes arduous when the robots are assigned different orbital radii, as their sensors or roles may demand. This paper presents a safety-critical control framework for this problem. Each robot commands both its acceleration and its turn-rate, measures only its distance to the target, and exchanges information only with its two ring neighbours. The design has three features. First, a structural property of the model lets one control law serve every robot regardless of its assigned radius. Second, the acceleration drives each radius exactly to its target, while the turn-rate balances the formation with the neighbours. Third, a higher-order control-barrier-function quadratic program enforces target, geofence, and inter-robot safety on both inputs at once; it leaves the nominal controller untouched away from the constraints and stays valid while the unmeasured range rate is still being estimated. Simulations confirm these guarantees. In a comparison, the closest existing multi-orbit design, even with full state information, violates both the target standoff and the inter-robot margin, whereas the proposed controller does not.
\end{abstract}


\section{Introduction}
\label{sec:intro}

\subsection{Motivation and Background}

\IEEEPARstart{C}{oordinated} circumnavigation of a stationary target
by autonomous robots underlies persistent surveillance of critical
infrastructure~\cite{shames2011circumnavigation}, monitoring of
hazardous sites~\cite{ranjan20253d}, offshore
inspection~\cite{paull2013auv}, standoff
reconnaissance~\cite{summers2009coordinated}, and oceanographic
sampling~\cite{fiorelli2007multi,lan2010distributed}. In each
setting the robots must converge to prescribed orbits, hold a
balanced formation with their neighbours, and never collide with the
target or with each other, at every instant and not merely in the
limit. The task is harder when robots are assigned distinct orbital
radii, as platforms differing in sensor resolution, payload, or role
naturally require: satellite constellations place high-resolution
imaging units inside relay units~\cite{petit2001constrained}, and
firefighting UAV teams place suppression platforms inside monitoring
platforms~\cite{merino2012unmanned}.

A single scalar input cannot meet these requirements: if only the
turn-rate is commanded, one degree of freedom must drive orbit
convergence, maintain safety, and coordinate the formation at once,
and these objectives compete for the same
channel~\cite{summers2009coordinated}. Commanding both acceleration
and turn-rate, the two-input model standard in collective
circular-motion studies~\cite{paley2007oscillator}, removes this
competition: the turn-rate serves the formation and the
acceleration serves radial convergence. Two difficulties remain.
First, the obstacle and inter-agent constraints have relative degree
two in the inputs, so hard, instantaneous safety requires a barrier
formulation solved jointly across both channels. Second, the target
is uncooperative, so each robot senses it through a single range
measurement, with no bearing, range rate, or absolute position; such
range-only sensors are also inexpensive, lightweight, and low-power,
making this a design choice as much as a
limitation~\cite{wang2024target,cao2015uav}. Teammates, by
contrast, are cooperative and exchange local information, so the
relative displacement to a neighbour is available. Any safety
certificate must therefore survive the error incurred while the
unmeasured range rate is being reconstructed. Addressing both
difficulties, together with heterogeneous orbit assignment, is the
subject of this paper.

\subsection{Related Work}

Stable circular orbits were first established for single unicycle
agents~\cite{summers2009coordinated,deghat2014localization} and
later extended to multi-agent
formations~\cite{oh2015survey,dou2020target}; cyclic
pursuit~\cite{marshall2004formations,marshall2006pursuit} supplied
foundational stability results for ring-coupled formations, and
circular formations for distributed source seeking under
communication constraints~\cite{brinon2015distributed} illustrate
the breadth of the problem. These designs, however, either decouple
orbit tracking from coordination without formal coupling
guarantees~\cite{deghat2014localization} or rely on potential fields
that carry no safety certificate~\cite{oh2015survey}. On
heterogeneous multi-orbit circumnavigation specifically, the closest
work~\cite{zheng2015distributed} uses virtual-agent normalization
with circulant-matrix diagonalization, but requires orbit-dependent
tuning and full velocity access and offers no hard safety guarantee;
the alternatives are time-varying switching
strategies~\cite{ceccarelli2008collective,lan2010distributed} or are
restricted to homogeneous radii~\cite{cao2016collective}.

Where safety is treated at all in this literature, it draws on
control barrier functions (CBFs)~\cite{wang2017safety}, barrier
certificates~\cite{panagou2015distributed}, or the asymmetric
barrier Lyapunov function (ABLF) of~\cite{tee2009barrier}. Our own
prior work certified hard geofencing for a single unicycle via a
M\"obius-transformed ABLF~\cite{11239434}, but that construction
folds safety into one tracking Lyapunov function and does not extend
to an obstacle constraint \emph{and} pairwise inter-agent
constraints acting jointly on two inputs~\cite{singh2025phase}. The
CBF-QP paradigm~\cite{ames2019control} is the natural fit instead:
it certifies forward invariance as a minimally invasive filter on
any nominal controller and has been applied to multi-robot collision
avoidance~\cite{borrmann2015control}, though not to heterogeneous
ring circumnavigation, and the relative-degree-two constraints of
our problem lie beyond standard CBF-QPs. The higher-order CBF
(HOCBF) construction of~\cite{xiao2019control} resolves exactly
this, and solving it jointly across acceleration and turn-rate is
the central technical device of this paper.

What remains is the sensing side. Range-only circumnavigation has
been studied for localization and
pursuit~\cite{shames2011circumnavigation} and for target
following~\cite{matveev2011range}, but these works pursue
convergence, not certified safety: to our knowledge, no existing
design makes a hard, instantaneous safety guarantee survive the
transient of the estimator that range-only sensing necessitates.
This is where the convergence-oriented and safety-oriented lines of
work fail to meet, and it is the gap this paper closes: a framework
combining two-input unicycle actuation, heterogeneous orbit
assignment, and hard obstacle and inter-agent safety enforced
through a jointly solved HOCBF-QP that remains valid under range-only
sensing of the target and local communication between neighbours.

\subsection{Main Contributions}

Against this backdrop, the paper makes three contributions:

\begin{itemize}
\item The body-frame radial error is shown to be algebraically
independent of the assigned orbital radius. As a result, one control
law serves every agent, whatever its orbit, and no orbit-dependent
normalization or tuning is needed.

\item A cascaded backstepping law in the
acceleration channel drives each orbital radius to its target
exactly and exponentially, rather than practically or in
time-average. The turn-rate channel is thereby freed for a
gradient-based formation law that uses only ring-neighbour data.

\item Safety is then delegated to a single HOCBF-QP. It enforces the obstacle, geofence, and
inter-agent constraints, all of relative degree two in the inputs. Because the range rate is unmeasured, a
super-twisting differentiator reconstructs it in finite time, and an
explicit constraint tightening keeps the safety guarantee valid
throughout the estimation transient, not only after it settles. In
simulation, the heterogeneous multi-orbit design
of~\cite{zheng2015distributed}, granted full state access, violates
both the target standoff and the inter-agent margin during its
transient, whereas the proposed controller keeps both margins
positive at every instant under range-only sensing.
\end{itemize}

\section{Preliminaries}
\label{sec:prelim}

\subsection{Notation}

The set of non-negative real numbers is denoted by $\mathbb{R}_+$,
the Euclidean norm by $\|\cdot\|$, and $(\cdot)^\top$ denotes the
transpose of a vector. A continuous, strictly increasing function
$\beta:\mathbb{R}_+\to\mathbb{R}_+$ with $\beta(0)=0$ and
$\beta(s)\to\infty$ as $s\to\infty$ is said to belong to class
$\mathcal{K}_\infty$~\cite{khalil2002nonlinear}. The signum function
is denoted by $\operatorname{sign}(\cdot)$, and $\mathbf{1}[\cdot]$
denotes the indicator function, equal to $1$ when its argument is
true and $0$ otherwise. For a differentiable $b:\mathbb R^n\to
\mathbb R$ and a vector field $f$, $L_fb\triangleq\nabla b^\top f$
denotes the Lie derivative of $b$ along $f$. Throughout, the
subscript $i$ indexes a robot and the subscript $ij$ a pair of
robots; a hat $(\hat\cdot)$ marks a quantity reconstructed from
measurements, a tilde $(\tilde\cdot)$ a tracking error, the
superscript $\mathrm o$ a reconstruction error, and the letter
$\epsilon$ a constant bound on a transient
quantity.

\subsection{Control Barrier Functions}

Consider a control-affine system
\begin{equation}
  \dot x = f(x) + g(x)v, \qquad x\in\mathbb R^n,\ v\in\mathcal U
  \subset\mathbb R^m,
  \label{eq:affine_sys}
\end{equation}
with $f,g$ locally Lipschitz and $v$ the control input. For a set
$\mathcal C\triangleq\{x\in\mathbb R^n : b(x)\ge0\}$ defined by a
continuously differentiable function $b:\mathbb R^n\to\mathbb R$,
$\mathcal C$ is \emph{forward invariant} under \eqref{eq:affine_sys}
if $x(0)\in\mathcal C\implies x(t)\in\mathcal C$ for all $t\ge0$.

\begin{definition}[Control Barrier Function~\cite{ames2019control}]
\label{defn:cbf}
The function $b$ is a control barrier function (CBF) for
\eqref{eq:affine_sys} on $\mathcal C$ if there exists a class
$\mathcal K_\infty$ function $\beta$ such that
\begin{equation}
  \sup_{v\in\mathcal U}\bigl[L_fb(x) + L_gb(x)v + \beta(b(x))\bigr]
  \ge 0, \qquad \forall\,x\in\mathcal C.
  \label{eq:cbf_condition}
\end{equation}
\end{definition}

\begin{corollary}[CBF Forward Invariance~\cite{ames2019control}]
\label{thm:cbf}
If $b$ is a CBF for \eqref{eq:affine_sys} on $\mathcal C$ and the
control $v(t)$ is chosen, for every $t\ge0$, to satisfy
\begin{equation}
  L_fb(x(t)) + L_gb(x(t))\,v(t) + \beta(b(x(t))) \ge 0,
  \label{eq:cbf_enforced}
\end{equation}
then $\mathcal C$ is forward invariant.
\end{corollary}

Condition~\eqref{eq:cbf_enforced} can be met by a choice of $v$ only
if $L_gb(x)\ne0$ wherever $L_fb(x)+\beta(b(x))<0$; that is, $b$
must have relative degree one with respect to $v$. When the
constraint is a function of position alone and the input acts at the
acceleration level, $L_gb\equiv0$ identically, the input never
appears in \eqref{eq:cbf_enforced}, and Definition~\ref{defn:cbf} is
inapplicable. The following generalization treats exactly this case.

\begin{definition}[Higher-Order Control Barrier
Function~\cite{xiao2019control}]
\label{defn:hocbf}
Let $b:\mathbb R^n\to\mathbb R$ be $r$-times continuously
differentiable with relative degree $r\ge2$ in $v$ for
\eqref{eq:affine_sys}. Given differentiable class $\mathcal K_\infty$
functions $\beta_1,\ldots,\beta_{r-1}$, define the sequence
\begin{equation}
  \psi_0(x) \triangleq b(x), \quad
  \psi_k(x) \triangleq \dot\psi_{k-1}(x) + \beta_k(\psi_{k-1}(x)),
  \label{eq:hocbf_chain}
\end{equation}
for $k=1,\ldots,r-1$, each $\psi_k$ a function of $x$ alone, since
the relative-degree assumption keeps $v$ out of the first $r-1$
derivatives of $b$, together with the nested sets
$\mathcal C_k\triangleq\{x: \psi_{k-1}(x)\ge0\}$, $k=1,\ldots,r$.
Then $b$ is a higher-order control barrier function (HOCBF) for
\eqref{eq:affine_sys} if there exists a class $\mathcal K_\infty$
function $\beta_r$ such that
\begin{equation}
  \sup_{v\in\mathcal U}\bigl[\dot\psi_{r-1}(x,v)
  + \beta_r(\psi_{r-1}(x))\bigr]
  \ge 0, \qquad \forall\,x\in\bigcap_{k=1}^{r}\mathcal C_k.
  \label{eq:hocbf_condition}
\end{equation}
\end{definition}

\begin{corollary}[HOCBF Forward Invariance~\cite{xiao2019control}]
\label{thm:hocbf}
If $b$ is a HOCBF for \eqref{eq:affine_sys} per
Definition~\ref{defn:hocbf}, $x(0)\in\bigcap_{k=1}^r\mathcal C_k$,
and the control $v(t)$ is chosen, for every $t\ge0$, to satisfy
\begin{equation}
  \dot\psi_{r-1}(x(t),v(t)) + \beta_r(\psi_{r-1}(x(t))) \ge 0,
  \label{eq:hocbf_enforced}
\end{equation}
then $x(t)\in\bigcap_{k=1}^{r}\mathcal C_k\subseteq\mathcal
C_1=\{x:b(x)\ge0\}$ for all $t\ge0$.
\end{corollary}

For $r=2$, the only case needed in this paper,
\eqref{eq:hocbf_chain}--\eqref{eq:hocbf_enforced} reduce to
\begin{align}
\nonumber  \psi_0(x)=b(x),\quad
  &\psi_1(x)=\dot b(x)+\beta_1(b(x)),\\
  &\dot\psi_1(x,v)+\beta_2(\psi_1(x))\ge0.
  \label{eq:hocbf_r2}
\end{align}
Verifying that a candidate $b$ is a valid HOCBF then reduces to
two checkable conditions:
\begin{enumerate}[label=\textnormal{(C\arabic*)}]
\item \label{cond:affine} $\ddot b(x,v)$ depends on $v$ and is
  affine in it, so that $b$ has relative degree exactly $2$ and
  \eqref{eq:hocbf_r2} yields a linear constraint on $v$;
\item \label{cond:nonvanish} the gain vector multiplying $v$ in
  $\dot\psi_1(x,v)$ is nonzero on the region of interest, so that
  \eqref{eq:hocbf_enforced} is solvable for some $v\in\mathcal U$.
\end{enumerate}
Both conditions are verified explicitly for each barrier
constructed in Section~\ref{sec:stability}. The next section
describes the system under consideration and formalizes the
control problem.

\section{System Model and Problem Formulation}
\label{sec:problem}

\subsection{Unicycle Kinematics}

We consider $N$ robots moving in the plane $\mathbb R^2$,
circumnavigating a stationary target fixed at the origin and treated
as a circular obstacle of radius $r_{\min}>0$. Robot
$i\in\{1,\ldots,N\}$ has position $p_i=[x_i,y_i]^\top\in\mathbb R^2$,
heading $\varphi_i\in[0,2\pi)$, and forward speed $u_i(t)$, evolving
as
\begin{equation}
  \dot{x}_i = u_i\cos\varphi_i, \quad
  \dot{y}_i = u_i\sin\varphi_i, \quad
  \dot{\varphi}_i = \omega_i, \quad \dot u_i = a_i,
  \label{eq:unicycle}
\end{equation}
where the acceleration $a_i(t)\in[a_{\min},a_{\max}]$ and turn-rate
$\omega_i(t)\in[-\omega_{\max},\omega_{\max}]$ are the two commanded
inputs, with $a_{\min}<0<a_{\max}$ and $\omega_{\max}>0$. This is
the standard second-order (dynamic) unicycle model, appropriate when
actuator commands are forces/torques rather than instantaneous
velocities~\cite{summers2009coordinated,paley2007oscillator,singh2025phase}.
Note that the forward speed $u_i$ is a state of
\eqref{eq:unicycle}, not an input. Associated with the heading,
define the unit vectors
\begin{equation}
  h_i \triangleq \begin{bmatrix}\cos\varphi_i\\ \sin\varphi_i
  \end{bmatrix}, \qquad
  \hpi \triangleq \begin{bmatrix}-\sin\varphi_i\\ \cos\varphi_i
  \end{bmatrix},
  \label{eq:heading_vectors}
\end{equation}
so that $\dot p_i = u_i h_i$ and $\dot h_i = \omega_i\hpi$.

\begin{figure}[t]
\centering
\includegraphics[width=0.9\linewidth]{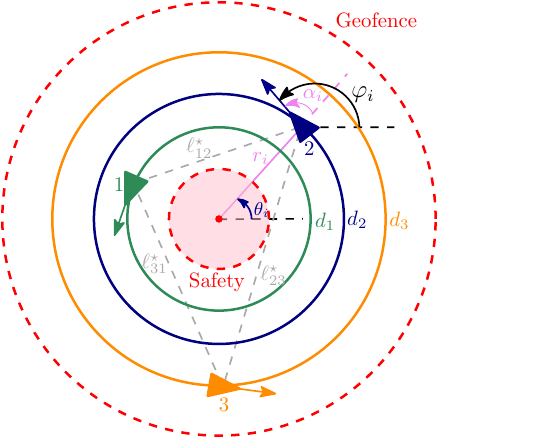}
\caption{Problem setup for $N=3$: robots on heterogeneous orbits
$d_1<d_2<d_3$ around a stationary target, confined to
the annulus between the safety boundary $r_{\min}$ and the
geofence $r_{\max}$, coordinating over the ring
graph with desired chord lengths $\ell_{ij}^\star$ (gray dashed).}
\label{fig:problem_setup}
\vspace{-0.25cm}
\end{figure}

With the range $r_i\triangleq\|p_i\|$, bearing
$\theta_i\triangleq\mathrm{atan2}(y_i,x_i)$, and heading offset
$\alpha_i\triangleq\varphi_i-\theta_i$ from the outward radial
direction, differentiating along \eqref{eq:unicycle} gives, for
$r_i>0$,
\begin{equation}
  \dot r_i = u_i\cos\alpha_i, \quad
  \dot\theta_i = \frac{u_i\sin\alpha_i}{r_i}, \quad
  \dot\alpha_i = \omega_i - \frac{u_i\sin\alpha_i}{r_i}.
  \label{eq:ralpha_dyn}
\end{equation}
Range $r_i$, bearing $\theta_i$, and heading offset $\alpha_i$
are depicted in Fig.~\ref{fig:problem_setup}. With
the individual robot dynamics fixed, we next specify how the $N$
robots are coordinated with each other.

\subsection{Interaction Topology}
\label{ssec:topology}

Robot $i$ is assigned a desired orbital radius $d_i>r_{\min}$, with
the rings ordered as $d_1<d_2<\cdots<d_N$. The robots coordinate
over a cyclic ring graph $\mathcal G=(\mathcal V,\mathcal E)$ with
$\mathcal V=\{1,\ldots,N\}$ and undirected edge set
\begin{equation}
  \mathcal{E} = \bigl\{\{i,\,i+1\} : i \in \mathcal{V}\bigr\},
  \label{eq:edges}
\end{equation}
node labels taken modulo $N$ (so that $N+1$ is identified with
$1$). Each robot thus has the two ring neighbours
$\mathcal N_i=\{i-1,\, i+1\}$, again modulo $N$. The only
team-level information robot $i$ requires a priori is the team
size $N$, its own radius $d_i$, and the radii $d_j$ of its two
ring neighbours $j\in\mathcal N_i$; no robot needs the radii or
positions of the remaining teammates. For $N=3$ the ring
coincides with the complete graph, whereas for $N\ge4$ it is
strictly sparser, each robot interacting with only two of its
$N-1$ teammates.

\subsection{Target Angular Rate and Formation Geometry}
\label{ssec:geometry}

Let $\omega_0>0$ denote a common phase-advance rate: at the target
formation, all robots orbit the origin at angular speed $\omega_0$
with equal angular separation $2\pi/N$ between ring-adjacent
robots. The desired configuration is thus
\begin{equation}
  p_i^\star(t) = d_i\begin{bmatrix}\cos\theta_i^\star(t)\\
  \sin\theta_i^\star(t)\end{bmatrix}, \qquad
  \theta_{i+1}^\star-\theta_i^\star=\frac{2\pi}{N},
  \label{eq:pstar}
\end{equation}
with $\dot\theta_i^\star=\omega_0$ for all $i$. For each edge
$\{i,j\}\in\mathcal E$, the corresponding desired chord length
$\ell_{ij}^\star$ follows from the law of cosines with included
angle $2\pi/N$:
\begin{equation}
  \ell_{ij}^\star = \sqrt{d_i^2 + d_j^2 - 2d_id_j\cos\!\left(
  \frac{2\pi}{N}\right)};
  \label{eq:chord}
\end{equation}
when $d_i\equiv d$, this reduces to the familiar
$\ell^\star=2d\sin(\pi/N)$. Further, \eqref{eq:chord} gives the
desired separation of \emph{ring-adjacent} pairs only; non-adjacent
pairs are separated at the target formation by
$\|p_i^\star-p_j^\star\|$ directly, with $p_i^\star$ as in
\eqref{eq:pstar}. The heterogeneous orbits and the corresponding
chords $\ell_{ij}^\star$ are shown in Fig.~\ref{fig:problem_setup}.
These chord lengths are the target inter-agent distances that the
safety margin introduced next must remain compatible with.

\subsection{Obstacle and Inter-Agent Safety}
\label{ssec:safety}

The hard safety requirements are confinement to an operating
annulus and inter-agent collision avoidance: for a geofence radius
$r_{\max}>d_N$ and a safety margin $d_{\mathrm{safe}}>0$,
\begin{equation}
  r_{\min} \le r_i(t) \le r_{\max} \ \ \forall\,i, \qquad
  \|p_i(t)-p_j(t)\| \ge d_{\mathrm{safe}} \ \ \forall\, i\ne j,
  \label{eq:safety}
\end{equation}
both holding at every instant $t\ge0$: the lower bound keeps every
robot clear of the target, and the upper bound confines it to the
mission area. Since $r_{\min}>0$, the lower bound also keeps
$r_i(t)>0$ at every instant, so the polar description
\eqref{eq:ralpha_dyn} remains well defined throughout.
Compatibility of $d_{\mathrm{safe}}$ with the formation geometry
requires
\begin{equation}
  d_{\mathrm{safe}} < \min_{\{i,j\}\in\mathcal E}\ell_{ij}^\star,
  \qquad
  d_{\mathrm{safe}} < \min_{\substack{i\ne j\\ \{i,j\}\notin\mathcal E}}
  \|p_i^\star-p_j^\star\|,
  \label{eq:dsafe_compat}
\end{equation}
where $p_i^\star$ is given by \eqref{eq:pstar}: the first condition
ensures the safety margin does not conflict with the desired chord
lengths, and the second that non-adjacent robots are separated by
more than $d_{\mathrm{safe}}$ at the target formation.

\begin{assumption}
\label{ass:init}
The initial condition satisfies $r_{\min}<r_i(0)\le r_{\max}$ for
all $i$, $\|p_i(0)-p_j(0)\|>d_{\mathrm{safe}}$ for all $i\ne j$,
and $u_i(0)\in[u_{\min},u_{\max}]$, where $0<u_{\min}<u_{\max}<\infty$
are design constants fixed a priori and enforced as a speed envelope
throughout.
\end{assumption}

\subsection{Sensing and Communication}
\label{ssec:sensing}

Each robot has proprioceptive access to its own heading
$\varphi_i(t)$ and forward speed $u_i(t)$, and knows the inputs
$(a_i,\omega_i)$ it applies. The two remaining channels, toward
teammates and toward the target, differ fundamentally in what they
can recover, and it is worth being explicit about why.

Between robots, sensing is \emph{cooperative} and is supported by
explicit local communication. Each robot carries a transponder and
broadcasts its odometry, so robot $i$ recovers, through onboard
relative localization (e.g.\ UWB ranging fused with
odometry~\cite{ze2026relative}, or vision-based detection of the
teammate), the full two-dimensional relative displacement
\begin{equation}
  z_{ij}\triangleq p_i-p_j
  \label{eq:zij_def}
\end{equation}
to a teammate $j$. Because the neighbour actively participates in
the measurement, both the distance and the direction to it are
obtained, that is, the entire vector $z_{ij}$ and not merely its
magnitude. Robot $i$ acquires $z_{ij}$ for its ring neighbours
$j\in\mathcal N_i$ at all times, and for any other robot $j$
whenever $\|z_{ij}\|\le R_s$, where $R_s>d_{\mathrm{safe}}$ is the
sensing and communication radius. The set of robots with which
robot $i$ exchanges information at time $t$ is therefore
\begin{equation}
  \mathcal A_i(t)\triangleq\mathcal N_i\cup
  \bigl\{j\neq i:\norm{z_{ij}(t)}\leq R_s\bigr\},
  \label{eq:active_set}
\end{equation}
and over the same link each $j\in\mathcal A_i(t)$ communicates
its current speed, heading, and applied inputs
$(u_j,\varphi_j,a_j,\omega_j)$ to robot $i$. No robot communicates,
or needs, its absolute position, which under the target-sensing
model below it does not possess. Robots farther apart than $R_s$
neither measure nor communicate with one another, but neither can
they collide without first crossing the annulus
$d_{\mathrm{safe}}<\|z_{ij}\|\le R_s$, in which mutual
measurements, and hence a mutual safety constraint, become
available.

Toward the target, sensing is different, because the target is
\emph{non-cooperative}: it carries no transponder and shares no
data, so the relative localization used between teammates does not
apply, and the relative-position vector $-p_i$ is not recoverable.
The only modality that works against an uninstrumented object is
bearing-agnostic ranging (radar, acoustic time-of-flight, or a
single-beam rangefinder), which returns distance but not direction.
Robot $i$ therefore measures only the scalar range $r_i(t)$, with
no bearing $\theta_i(t)$, no range rate $\dot r_i(t)$, and no
absolute position $p_i(t)$: the \emph{range-only} sensing model of
distance-based circumnavigation~\cite{shames2011circumnavigation}
and range-only target following~\cite{matveev2011range}. Since
such sensors are also inexpensive, lightweight, and low-power, the
model is a realistic design point rather than an imposed
limitation.

\subsection{Problem Statement}

\begin{problem}
\label{prob:main}
Under Assumption~\ref{ass:init} with the sensing and communication
model, and for the setup of
Fig.~\ref{fig:problem_setup}, design $a_i(t)\in[a_{\min},a_{\max}]$
and $\omega_i(t)\in[-\omega_{\max},\omega_{\max}]$ for each robot
$i$, using only $r_i(t)$, $u_i(t)$, $\varphi_i(t)$, the measured
relative displacements $z_{ij}(t)$, and the communicated
$(u_j,\varphi_j,a_j,\omega_j)$ for $j\in\mathcal A_i(t)$, to ensure:
\begin{penumerate}
  \item \textit{Orbit and ring-formation convergence:}
    $r_i(t)\to d_i$ for all $i$, and
    $\|p_i(t)-p_j(t)\|\to\ell_{ij}^\star$ for all
    $\{i,j\}\in\mathcal E$, as $t\to\infty$;
  \item \textit{Annulus confinement:} $r_{\min}\le r_i(t)\le r_{\max}$
    for all $t\ge0$ and all $i$;
  \item \textit{Inter-agent collision avoidance:}
    $\|p_i(t)-p_j(t)\|\ge d_{\mathrm{safe}}$ for all $t\ge0$ and all
    $i\ne j$.
\end{penumerate}
\end{problem}

\section{Control Design}
\label{sec:control}

\begin{figure}[t]
\centering
\begin{tikzpicture}[
  font=\scriptsize,
  blk/.style={draw,rounded corners=2pt,align=center,minimum height=9mm,
              minimum width=22mm,inner sep=2pt},
  est/.style={blk,fill=blue!8},
  ctl/.style={blk,fill=orange!12},
  qpb/.style={blk,fill=green!12,thick},
  pl/.style={blk,fill=gray!12},
  arr/.style={-{Latex[length=1.5mm]},semithick},
  lab/.style={font=\scriptsize,inner sep=1.5pt,fill=white},
  lab2/.style={font=\scriptsize,inner sep=1pt}]
\node[pl]  (rob) at (0,0)      {Robot $i$\\unicycle \eqref{eq:unicycle}};
\node[est] (sta) at (3.1,0)    {Super-twisting\\differentiator \eqref{eq:sta1}--\eqref{eq:sta2}};
\node[est] (rec) at (6.2,0)    {Algebraic\\reconstruction \eqref{eq:ex_hat}, \eqref{eq:ey_hat}};
\node[ctl] (ph)  at (0,-2.1)   {Ring-formation\\phase law \eqref{eq:phase_law}};
\node[qpb] (qpn) at (3.1,-2.1) {HOCBF-QP\\safety filter \eqref{eq:QP}};
\node[ctl] (bs)  at (6.2,-2.1) {Backstepping\\acceleration law \eqref{eq:a_ctrl}};
\node[align=center] (nb) at (1.55,-3.5)
  {$z_{ij}$, $(u_j,\varphi_j,a_j,\omega_j)$ from neighbours $j\in\mathcal A_i(t)$};
\draw[arr] (rob) -- node[lab2,above=1pt]{$r_i$} (sta);
\draw[arr] (sta) -- node[lab2,above=1pt]{$\hat{\dot r}_i$} (rec);
\draw[arr] (rec) -- node[lab,right]{$\hat{\bar e}_{x,i},\,\hat{\bar e}_{y,i}$} (bs);
\draw[arr] (bs)  -- node[lab2,above=1pt]{$a_i^{\mathrm{nom}}$} (qpn);
\draw[arr] (ph)  -- node[lab2,above=1pt]{$\omega_i^{\mathrm{nom}}$} (qpn);
\draw[arr] (qpn.north) -- ++(0,0.5)
  -| node[lab,pos=0.25,above]{$a_i^\star,\,\omega_i^\star$}
  ([xshift=5mm]rob.south);
\draw[arr] ([xshift=-5mm]rob.south) -- node[lab,left]{$\varphi_i$}
  ([xshift=-5mm]ph.north);
\draw[arr] (nb.north -| ph)  -- (ph.south);
\draw[arr] (nb.north -| qpn) -- (qpn.south);
\end{tikzpicture}
\caption{Control architecture of robot $i$. Estimation blocks (blue)
use only the measured range $r_i$; the nominal controller (orange)
produces $(a_i^{\mathrm{nom}},\omega_i^{\mathrm{nom}})$; the
HOCBF-QP filter (green) enforces the safety constraints
\eqref{eq:hocbf_obs}, \eqref{eq:hocbf_geo}, \eqref{eq:speed_cbf},
\eqref{eq:hocbf_int} on both inputs jointly.}
\label{fig:block_diagram}
\vspace{-0.25cm}
\end{figure}
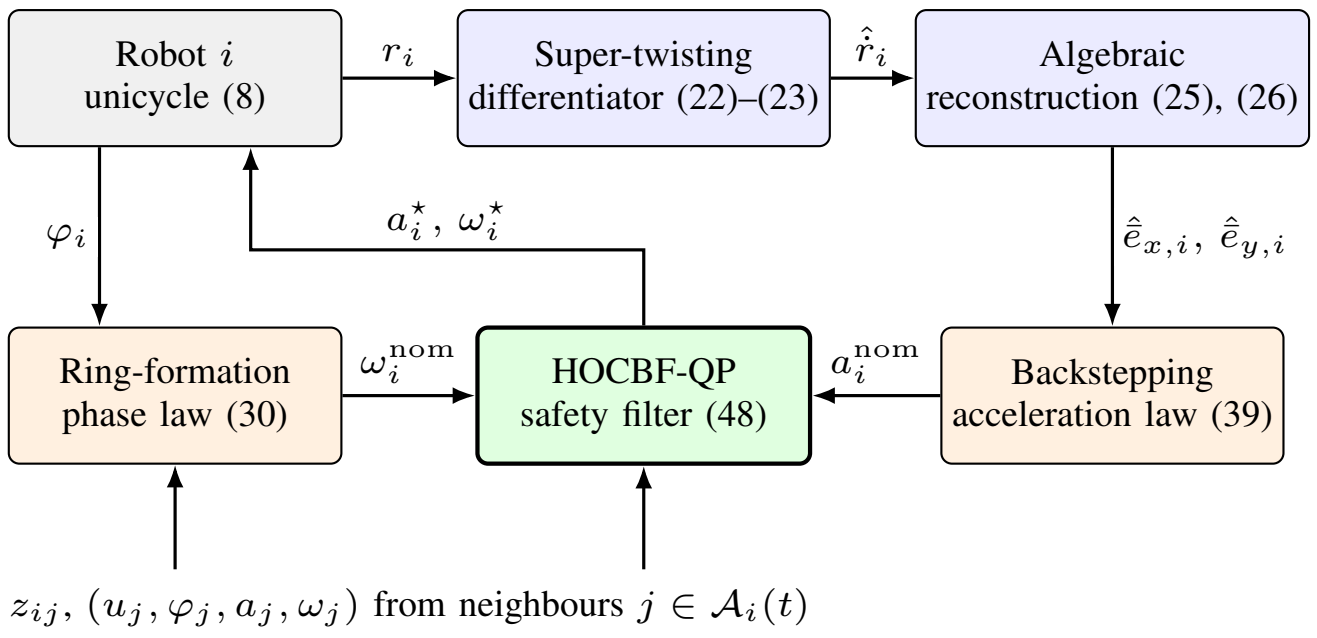

The design is layered, as Fig.~\ref{fig:block_diagram} summarizes.
This section builds the nominal inputs $\omega_i^{\mathrm{nom}}$
and $a_i^{\mathrm{nom}}$ that achieve the convergence objective
(P1) from range-only measurements. The safety filter of the next
section then takes these as its target and alters them only when a
safety constraint is active. The filter is designed so that,
whatever the nominal law does, it maintains for all $t\ge0$:
\begin{enumerate}[label=\textnormal{(F\arabic*)},leftmargin=*]
\item \label{fact:speed} the speed envelope
  $u_i(t)\in[u_{\min},u_{\max}]$;
\item \label{fact:range} the operating annulus
  $r_{\min}\le r_i(t)\le r_{\max}$ of \eqref{eq:safety}.
\end{enumerate}
These are the only properties of the filter used here, so the
radial objective below is a plain tracking error with no barrier
structure of its own. The signal flow is simple: the measured range
$r_i$ feeds a super-twisting differentiator, whose output is used
to reconstruct the body-frame coordinates algebraically; these
drive a backstepping acceleration law, while the ring-neighbour
data drive the turn-rate through a formation phase law; both
nominal inputs then pass through the HOCBF-QP filter.

\subsection{Body-Frame Coordinates and the Radial
Identity}
\label{ssec:transform}

The convergence objective (P1) is to drive each range $r_i$ to its
assigned radius $d_i$, so the primary regulated quantity is the
radial error $e_{d,i}\triangleq r_i-d_i$, whose vanishing places
robot $i$ exactly on its orbit. Following the standard approach in
circumnavigation control~\cite{summers2009coordinated,dou2020target},
introduce for each robot $i$ the transformed body-frame coordinates
\begin{equation}
  \begin{bmatrix}\bar{e}_{x,i}\\\bar{e}_{y,i}\end{bmatrix}
  =
  \begin{bmatrix}\cos\varphi_i & \sin\varphi_i\\
    -\sin\varphi_i & \cos\varphi_i\end{bmatrix}
  p_i
  + d_i\begin{bmatrix}0\\1\end{bmatrix}.
  \label{eq:transform}
\end{equation}
In terms of the heading vectors \eqref{eq:heading_vectors} and the
heading offset $\alpha_i$ of \eqref{eq:ralpha_dyn},
\begin{equation}
  \bar e_{x,i}=h_i^\top p_i=r_i\cos\alpha_i,\qquad
  \bar e_{y,i}-d_i=\hpi^\top p_i=-r_i\sin\alpha_i,
  \label{eq:e_geometric}
\end{equation}
so $r_i^2=\bar e_{x,i}^2+(\bar e_{y,i}-d_i)^2$, and
$\bar e_{x,i}=\bar e_{y,i}=0$ exactly when $r_i=d_i$ and
$\alpha_i=\pi/2$ (tangential, counterclockwise motion on the target
orbit). By \eqref{eq:e_geometric}, $\bar e_{x,i}$ depends only on
$p_i$ and $\varphi_i$, not on $d_i$, so the regulation task
$\bar e_{x,i}\to0$ is identical in form for every robot, regardless
of its assigned radius. Differentiating \eqref{eq:transform} along
\eqref{eq:unicycle} gives
\begin{equation}
  \dot{\bar{e}}_{x,i} = u_i + (\bar{e}_{y,i} - d_i)\omega_i,
  \qquad
  \dot{\bar{e}}_{y,i} = -\bar{e}_{x,i}\omega_i,
  \label{eq:edyn}
\end{equation}
and combining \eqref{eq:e_geometric} with $\dot r_i=u_i\cos\alpha_i$
from \eqref{eq:ralpha_dyn} leads to the fundamental radial identity
\begin{equation}
  r_i\dot{r}_i = u_i\,\bar{e}_{x,i}.
  \label{eq:fund}
\end{equation}
With \ref{fact:speed} in force, \eqref{eq:fund} shows that driving
$\bar e_{x,i}\to0$ drives $\dot r_i\to0$: regulating the radial error
$e_{d,i}$ thus reduces to regulating $\bar e_{x,i}$. Under the
range-only sensing model, however,
neither $\bar e_{x,i}$ nor $\bar e_{y,i}$ is measured, both being
functions of the unavailable $p_i$, so they must first be
reconstructed from $r_i(t)$, $u_i(t)$, and $\varphi_i(t)$.

\subsection{Range-Rate Reconstruction via Super-Twisting
Differentiation}
\label{ssec:differentiator}

Since only $r_i(t)$ is measured, $\dot r_i(t)$ is reconstructed with
a super-twisting
differentiator~\cite{levant1998robust,moreno2012strict} driven by
$r_i(t)$ alone, with internal states $\zeta_{i,1},\zeta_{i,2}$:
\begin{align}
  \dot \zeta_{i,1}&=\zeta_{i,2}-\lambda_1\abs{\zeta_{i,1}-r_i}^{1/2}
  \mathrm{sign}(\zeta_{i,1}-r_i),
  \label{eq:sta1}\\
  \dot \zeta_{i,2}&=-\lambda_2\,\mathrm{sign}(\zeta_{i,1}-r_i),
  \label{eq:sta2}
\end{align}
with $\hat{\dot r}_i\triangleq\zeta_{i,2}$ the range-rate estimate and
estimation errors $\eta_{i,1}\triangleq\zeta_{i,1}-r_i$,
$\eta_{i,2}\triangleq\zeta_{i,2}-\dot r_i$.

\begin{assumption}[Bounded Radial Acceleration]
\label{ass:jerk}
There exists a known $L>0$ such that $\abs{\ddot r_i(t)}\leq L$ for
all $t\geq0$ and all $i$.
\end{assumption}

Assumption~\ref{ass:jerk} is not an independent hypothesis: from
\eqref{eq:ralpha_dyn}, $\ddot r_i=a_i\cos\alpha_i-
u_i\sin\alpha_i\,\dot\alpha_i$ with
$\dot\alpha_i=\omega_i-u_i\sin\alpha_i/r_i$, so, using the input
bounds and \ref{fact:speed}--\ref{fact:range},
\begin{equation}
  L\triangleq\max(\abs{a_{\min}},a_{\max})
  +u_{\max}\Bigl(\omega_{\max}+\frac{u_{\max}}{r_{\min}}\Bigr)
  \label{eq:L_def}
\end{equation}
is an explicit bound computable from quantities already fixed.

\begin{theorem}[Finite-Time Exact
Differentiation~\cite{levant1998robust,moreno2012strict}]
\label{thm:sta}
Under Assumption~\ref{ass:jerk}, if $\lambda_1,\lambda_2>0$ are
chosen per the Lyapunov-based tuning rule of~\cite{moreno2012strict}
for the bound $L$, then there exists $T_i^*<\infty$, depending only
on $L,\lambda_1,\lambda_2$ and the initial error, such that
$\eta_{i,1}(t)=\eta_{i,2}(t)=0$ for all $t\geq T_i^*$; that is,
$\zeta_{i,1}(t)=r_i(t)$ and $\hat{\dot r}_i(t)=\dot r_i(t)$ exactly
for $t\geq T_i^*$. Moreover, there is a computable $\epsilon_2<\infty$
with $\abs{\eta_{i,2}(t)}\leq\epsilon_2$ for all $t\in[0,T_i^*)$.
\end{theorem}

\begin{remark}[Computable Convergence Time and Online Detection]
\label{rem:Tstar}
Initializing $\zeta_{i,1}(0)=r_i(0)$ (measured) and
$\zeta_{i,2}(0)=0$ gives $\eta_{i,1}(0)=0$ and
$\abs{\eta_{i,2}(0)}=\abs{\dot r_i(0)}\leq u_{\max}$ by
\ref{fact:speed}, so the strict Lyapunov function
of~\cite{moreno2012strict} yields both the transient bound
$\epsilon_2$ and a computable upper bound
$\bar T\geq\max_iT_i^*$ from $(L,\lambda_1,\lambda_2,u_{\max})$
alone. In implementation, convergence can equivalently be detected
online, as the onset of the sliding condition $\zeta_{i,1}\equiv r_i$
sustained over a dwell window. Either mechanism provides a causal
switching signal; $T_i^*$ itself is never required in advance. We
write $T^*\triangleq\max_iT_i^*\le\bar T$.
\end{remark}

\subsection{Range-Only Reconstruction of Body-Frame Coordinates}
\label{ssec:reconstruction}

Substituting $\hat{\dot r}_i$ for $\dot r_i$ in \eqref{eq:fund}
defines, using only measured/estimated quantities,
\begin{equation}
  \hat{\bar e}_{x,i}\triangleq\frac{r_i\,\hat{\dot r}_i}{u_i},
  \qquad
  \tilde e_{x,i}^{\mathrm{o}}\triangleq\hat{\bar e}_{x,i}-\bar e_{x,i}
  =\frac{r_i}{u_i}\eta_{i,2}.
  \label{eq:ex_hat}
\end{equation}
By Theorem~\ref{thm:sta} and \ref{fact:speed}--\ref{fact:range},
$\tilde e_{x,i}^{\mathrm o}(t)=0$ for $t\geq T_i^*$, and
$\abs{\tilde e_{x,i}^{\mathrm o}(t)}\leq\epsilon_x\triangleq
r_{\max}\epsilon_2/u_{\min}$ for $t<T_i^*$.

The remaining coordinate is reconstructed \emph{algebraically}, not
by integrating \eqref{eq:edyn}: an integrator does not forget a
transient input, so integration would leave a permanent bias even
after $\tilde e_{x,i}^{\mathrm o}\to0$. Instead, from
$r_i^2=\bar e_{x,i}^2+(\bar e_{y,i}-d_i)^2$,
\begin{equation}
  \hat{\bar e}_{y,i}\triangleq d_i+s_i\sqrt{\max\bigl(r_i^2-
  \hat{\bar e}_{x,i}^2,\,0\bigr)},
  \label{eq:ey_hat}
\end{equation}
where, by \eqref{eq:e_geometric}, the correct sign is
$s_i=-\operatorname{sign}(\sin\alpha_i)$. The initial value
$s_i(0)$ is assumed known at deployment (it only encodes whether the
robot initially moves with clockwise or counterclockwise tendency
about the target), and $s_i$ is updated at the isolated instants
$\hat{\bar e}_{x,i}=\pm r_i$, where $\bar e_{y,i}=d_i$, i.e.\
purely radial motion, tracked by continuity of
$\hat{\bar e}_{y,i}$ across the crossing. Because \eqref{eq:ey_hat}
is a static (memoryless) function of $\hat{\bar e}_{x,i}$ and
$r_i$, exactness is inherited instantly: for $t\geq T^*$,
$\hat{\bar e}_{y,i}(t)=\bar e_{y,i}(t)$, with no accumulated bias.
For $t<T^*$, the error $\tilde e_{y,i}^{\mathrm o}\triangleq
\hat{\bar e}_{y,i}-\bar e_{y,i}$ is bounded in terms of $\epsilon_x$:
with $s_i$ correct, $\abs{\tilde e_{y,i}^{\mathrm o}}=
\bigl|\sqrt{\max(r_i^2-\hat{\bar e}_{x,i}^2,0)}-
\sqrt{r_i^2-\bar e_{x,i}^2}\bigr|$, and since
$\abs{\sqrt A-\sqrt B}\le\sqrt{\abs{A-B}}$ for $A,B\ge0$ (the
clipping at zero only moves the first argument toward the second),
\begin{equation}
  \abs{\tilde e_{y,i}^{\mathrm o}(t)}\le
  \sqrt{\abs{\hat{\bar e}_{x,i}^2-\bar e_{x,i}^2}}
  \le\epsilon_y\triangleq
  \sqrt{\epsilon_x\,(2r_{\max}+\epsilon_x)},
  \label{eq:ey_bound}
\end{equation}
using $\abs{\hat{\bar e}_{x,i}+\bar e_{x,i}}\le2r_{\max}+\epsilon_x$
by \eqref{eq:e_geometric} and \ref{fact:range}, uniformly in $i$.

\subsection{Ring Formation Potential and the Phase Law}
\label{ssec:formation}

With $z_{ij}$ as in \eqref{eq:zij_def} and $\ell_{ij}^\star$ from
\eqref{eq:chord}, define for each edge the squared-distance error
and, for the stacked bearing vector
$\theta\triangleq[\theta_1,\ldots,\theta_N]^\top$, the formation
potential
\begin{equation}
  e_{ij} \triangleq \|z_{ij}\|^2 - (\ell_{ij}^\star)^2, \qquad
  \Phi(\theta) \triangleq \frac{1}{2}
  \sum_{\{i,j\}\in\mathcal E}e_{ij}(\theta)^2,
  \label{eq:Vf}
\end{equation}
the latter evaluated on the target-radius manifold
$p_i=d_i[\cos\theta_i,\sin\theta_i]^\top$. At a balanced
configuration $\theta^\star$, a critical point of $\Phi$ with
$e_{ij}(\theta^\star)=0$ on every edge, the Hessian reduces to
$\nabla^2\Phi(\theta^\star)=\sum_{\{i,j\}\in\mathcal E}
(\nabla e_{ij})(\nabla e_{ij})^\top$, a sum of rank-one positive
semidefinite terms; it is positive definite on the subspace
transverse to uniform rotation whenever the edge gradients span
that subspace, which is what local exponential stability of
$\theta^\star$ for a gradient flow requires. For $N=3$ the ring is
the complete graph and the balanced configuration is, up to
rotation and reflection, the only configuration with
$e_{ij}=0$ on every edge; for $N\ge4$, as for any distance-based
potential on a sparse graph, $\Phi$ admits further critical points
(e.g.\ configurations in which the ring folds back on itself), so
the formation guarantee stated in Section~\ref{sec:stability} is
local to $\theta^\star$. Everything in \eqref{eq:Vf} uses only
$z_{ij}$, available through the ring-neighbour localization, and is
entirely independent of the target-range reconstruction above.

Let $S_i\triangleq\sum_{j\in\mathcal N_i} z_{ij}e_{ij}$, so
that $\nabla_{p_i}\Phi=2S_i$. Define the world-frame tangent
direction at $p_i$,
\begin{equation}
  \tilde h_{\perp,i}\triangleq
  \begin{bmatrix}-\sin\theta_i\\ \cos\theta_i\end{bmatrix}
  =\frac1{r_i}\begin{bmatrix}-y_i\\ x_i\end{bmatrix};
  \label{eq:htilde_def}
\end{equation}
on the target-radius manifold, $\partial p_i/\partial\theta_i=
d_i\tilde h_{\perp,i}$ and hence $\partial\Phi/\partial\theta_i=
2d_i\tilde h_{\perp,i}^\top S_i$. A gradient-descent phase law would
therefore steer against $\tilde h_{\perp,i}^\top S_i$; however,
$\tilde h_{\perp,i}$ is built from the bearing $\theta_i$, which is
unmeasured. We assign instead, in terms of the measured heading
vector $h_i$ of \eqref{eq:heading_vectors} and a formation gain
$\kappa_w>0$,
\begin{equation}
  \omega_i^{\mathrm{nom}}(t) \triangleq \omega_0
  - \kappa_w\, h_i^\top S_i(t),
  \label{eq:phase_law}
\end{equation}
computable from $\varphi_i(t)$ and the ring-neighbour data
$z_{ij}(t)$ alone, with no dependence on $\theta_i$, $r_i$, or the
range rate; it is applied as-is, unaffected by the differentiator. On the tangential manifold
$\alpha_i\equiv\pi/2$, i.e.\ $\varphi_i=\theta_i+\pi/2$, one has
$h_i=\tilde h_{\perp,i}$ exactly, so \eqref{eq:phase_law} coincides
there with the gradient law motivating it; off that manifold the two
differ.

\begin{remark}[Scope of the Descent Claim for $\Phi$]
\label{rem:phase_scope}
Two gaps separate \eqref{eq:phase_law} from the descent identity
$\dot\Phi=-2\kappa_w\sum_id_i(\tilde h_{\perp,i}^\top S_i)^2\le0$
motivating it. First, \eqref{eq:phase_law} uses the measured $h_i$
in place of $\tilde h_{\perp,i}$; the two agree only on the
tangential manifold $\alpha_i\equiv\pi/2$. Second, and more
fundamentally, the identity would require
$\dot\theta_i\equiv\omega_i^{\mathrm{nom}}$, whereas from
\eqref{eq:ralpha_dyn} the true bearing rate is $\dot\theta_i=
u_i\sin\alpha_i/r_i$: the turn-rate enters only indirectly, through
$\dot\alpha_i$, and matches $\dot\theta_i$ only on the same manifold
$\alpha_i\equiv\pi/2$ with, in addition,
$u_i=d_i\omega_i^{\mathrm{nom}}$ pointwise, a matching condition
not separately imposed. Consequently, $\dot\Phi\le0$ is to be read
as the behaviour on the reduced manifold $\alpha_i\equiv\pi/2$,
$r_i\equiv d_i$, which the radial cascade constructed below reaches
asymptotically and on which both gaps close simultaneously, not as
an identity on the full state space.
\end{remark}

\subsection{Cascaded Backstepping Error Hierarchy}
\label{ssec:backstep_setup}

With $\omega_i^{\mathrm{nom}}$ fixed by \eqref{eq:phase_law}, define
the computable feedforward
\begin{equation}
  c_i(t) \triangleq
  (\hat{\bar e}_{y,i}(t)-d_i)\,\omega_i^{\mathrm{nom}}(t),
  \label{eq:ci_def}
\end{equation}
built from the reconstruction $\hat{\bar e}_{y,i}$, since
$\bar e_{y,i}$ itself is unmeasured. When the applied turn-rate
equals $\omega_i^{\mathrm{nom}}$, substituting \eqref{eq:ci_def}
into \eqref{eq:edyn} gives
\begin{equation}
  \dot{\bar e}_{x,i}=u_i+c_i(t)
  -\tilde e_{y,i}^{\mathrm o}\,\omega_i^{\mathrm{nom}},
  \label{eq:exdyn_ci}
\end{equation}
with the coefficient of $u_i$ exactly $1$ at every state, and a
correction term vanishing for $t\geq T^*$. This drives the three-stage
construction below.

\textbf{Stage 1 (radial error).} Take the Lyapunov function
$V_{r,i}\triangleq\tfrac12e_{d,i}^2$ and the virtual reference
\begin{equation}
  \bar e_{x,i}^\star \triangleq -\kappa_1 r_i e_{d,i}
  =-\kappa_1(d_i+e_{d,i})\,e_{d,i},
  \qquad
  \hat{\tilde e}_{x,i}\triangleq\hat{\bar e}_{x,i}
  -\bar e_{x,i}^\star,
  \label{eq:virtual_ref}
\end{equation}
with gain $\kappa_1>0$; note $\bar e_{x,i}^\star$ is a function of
$e_{d,i}$ alone. Writing $\bar e_{x,i}=\hat{\tilde e}_{x,i}+
\bar e_{x,i}^\star-\tilde e_{x,i}^{\mathrm o}$ and substituting into
the \emph{true} identity \eqref{eq:fund} gives, exactly and for all
$t\ge0$,
\begin{align}
 \nonumber \dot e_{d,i} &= -\kappa_1 u_i e_{d,i}
  + \frac{u_i}{r_i}\bigl(\hat{\tilde e}_{x,i}
  -\tilde e_{x,i}^{\mathrm o}\bigr),\\
  \dot V_{r,i} &= -\kappa_1 u_i e_{d,i}^2
  + \frac{u_i e_{d,i}}{r_i}\bigl(\hat{\tilde e}_{x,i}
  -\tilde e_{x,i}^{\mathrm o}\bigr).
  \label{eq:ed_stage1}
\end{align}
The decay rate is $\kappa_1u_i$, not a fixed $\kappa_1$: the radial
channel is speed-dependent, but still exponential, since
$u_i\geq u_{\min}>0$ by \ref{fact:speed}. Further,
\eqref{eq:ed_stage1} is not an approximation valid only after the
differentiator converges; only the size of
$\tilde e_{x,i}^{\mathrm o}$ changes across $t=T^*$, from bounded by
$\epsilon_x$ to identically zero.

\textbf{Stage 2 (transverse error).} Since $u_i$ is itself a state
driven by $a_i$, $\hat{\tilde e}_{x,i}$ cannot be driven to zero
directly. Define the second virtual reference
\begin{equation}
  u_i^\star(t) \triangleq -\kappa_2\hat{\tilde e}_{x,i}
  - c_i(t)
  + \frac{d\bar e_{x,i}^\star}{de_{d,i}}\dot e_{d,i},
  \qquad \kappa_2>0,
  \label{eq:u_star}
\end{equation}
and $\tilde u_i\triangleq u_i-u_i^\star$. This choice is exactly
what cancels the Stage-1 cross term: for $t\ge T^*$, where the
correction term in \eqref{eq:exdyn_ci} vanishes and
$\hat{\tilde e}_{x,i}$ coincides with the true
$\tilde e_{x,i}\triangleq\bar e_{x,i}-\bar e_{x,i}^\star$, and with
the applied turn-rate equal to $\omega_i^{\mathrm{nom}}$,
\begin{equation}
  \dot{\hat{\tilde e}}_{x,i}
  =\dot{\bar e}_{x,i}
  -\frac{d\bar e_{x,i}^\star}{de_{d,i}}\dot e_{d,i}
  =u_i-u_i^\star-\kappa_2\hat{\tilde e}_{x,i}
  =\tilde u_i-\kappa_2\hat{\tilde e}_{x,i},
  \label{eq:etilde_x_dyn}
\end{equation}
using \eqref{eq:u_star} in the last step; the term
$(d\bar e_{x,i}^\star/de_{d,i})\dot e_{d,i}$ already carries the
full dependence of $\bar e_{x,i}^\star$ on $e_{d,i}$ through the
chain rule, so no further radial correction is needed.

\textbf{Stage 3 (speed error).} The speed tracks its reference only
through $\dot u_i=a_i$, giving the third error
$\tilde u_i=u_i-u_i^\star$ with
$\dot{\tilde u}_i=a_i-\dot u_i^\star$, the coefficient of $a_i$
exactly $1$.

\begin{remark}[Why Stages 2--3 Are Not Claimed Exact on $[0,T^*)$]
\label{rem:no_exact_claim}
Identity \eqref{eq:etilde_x_dyn} holds for $t\ge T^*$; it does not
extend to $[0,T^*)$, where $\hat{\tilde e}_{x,i}$ is built from the
estimator state $\zeta_{i,2}$ and differs from the true transverse
error. The analysis in Section~\ref{sec:stability} therefore argues
in two phases: (a) on $[0,T^*)$, all three cascade errors remain
\emph{bounded}, not necessarily decaying, as a consequence of the
inputs being confined to their compact box and of
\ref{fact:speed}--\ref{fact:range}, with no claim about the
derivative of $\hat{\tilde e}_{x,i}$; and (b) for $t\geq T^*$, the
reconstruction is exact, so the closed-loop
vector field is \emph{identical} to the full-information cascade,
started from whatever bounded state phase (a) reached, at which
point a standard exponential Lyapunov argument applies unmodified.
\end{remark}

This produces a three-block one-way cascade
$e_{d,i}\to\hat{\tilde e}_{x,i}\to\tilde u_i$, driven at the bottom
by $a_i$, with the phase subsystem \eqref{eq:phase_law} evolving in
parallel (subject to Remark~\ref{rem:phase_scope}) and with
reconstruction error entering only through $\hat{\tilde e}_{x,i}$
and $c_i(t)$, and only on $[0,T^*)$. It remains to derive the
acceleration law closing the bottom of the cascade.

\subsection{Nominal Acceleration Law}
\label{ssec:accel}

The controller can only use quantities it actually has, $e_{d,i}$,
$\hat{\tilde e}_{x,i}$, $u_i$, $c_i(t)$, never the reconstruction
error $\tilde e_{x,i}^{\mathrm o}$, which is defined against the
true, unmeasured $\bar e_{x,i}$. Substituting the Stage-1 rate
$\dot e_{d,i}\to(u_i/r_i)\hat{\bar e}_{x,i}=
u_i[-\kappa_1e_{d,i}+\hat{\tilde e}_{x,i}/r_i]$ into
\eqref{eq:u_star}, the \emph{implemented} speed reference is
\begin{align}
\nonumber  u_i^\star = \hat F(e_{d,i},\hat{\tilde e}_{x,i},u_i,c_i)
  &\triangleq -\kappa_2\hat{\tilde e}_{x,i}-c_i(t)\\
  &-\kappa_1(r_i+e_{d,i})\,u_i\Bigl[-\kappa_1e_{d,i}
  +\frac{\hat{\tilde e}_{x,i}}{r_i}\Bigr],
  \label{eq:F_def}
\end{align}
using $d\bar e_{x,i}^\star/de_{d,i}=-\kappa_1(r_i+e_{d,i})$ from
\eqref{eq:virtual_ref}. Note that $\hat F$ contains no
$\tilde e_{x,i}^{\mathrm o}$ term by construction: the mismatch
between $\hat F$ and the full-information reference is exactly the
$\tilde e_{x,i}^{\mathrm o}$ term already carried in the true
dynamics \eqref{eq:ed_stage1}, not a separate quantity the
controller must know. Treating
$(e_{d,i},\hat{\tilde e}_{x,i},u_i,c_i)$ as the arguments of
$\hat F$,
\begin{equation}
  \frac{\partial\hat F}{\partial u_i}
  =-\kappa_1(r_i+e_{d,i})\Bigl[-\kappa_1e_{d,i}
  +\frac{\hat{\tilde e}_{x,i}}{r_i}\Bigr].
  \label{eq:dF_du}
\end{equation}

Imposing $\dot{\tilde u}_i=a_i^{\mathrm{nom}}-\dot u_i^\star=
-\kappa_3\tilde u_i$, $\kappa_3>0$, and expanding $\dot u_i^\star$
by the chain rule with the term $(\partial\hat F/\partial u_i)\,a_i$
isolated on the left, gives
\begin{equation}
  a_i^{\mathrm{nom}}=\frac{\hat G(\cdot)-\kappa_3\tilde u_i}
  {1-\partial\hat F/\partial u_i},
  \label{eq:a_ctrl}
\end{equation}
where $\hat G$ collects the remaining chain-rule terms
$(\partial\hat F/\partial e_{d,i})\,\dot e_{d,i}+
(\partial\hat F/\partial\hat{\tilde e}_{x,i})\,
\dot{\hat{\tilde e}}_{x,i}-\dot c_i$. Crucially, every rate
appearing in $\hat G$ is evaluated by \emph{model substitution}
rather than by differentiating estimator formulas:
\begin{equation}
  \dot e_{d,i}\to\hat{\dot r}_i,\quad
  \dot{\hat{\bar e}}_{x,i}\to u_i
  +(\hat{\bar e}_{y,i}-d_i)\,\omega_i^{\mathrm{nom}},\quad
  \dot{\hat{\bar e}}_{y,i}\to
  -\hat{\bar e}_{x,i}\,\omega_i^{\mathrm{nom}},
  \label{eq:model_subst}
\end{equation}
i.e., the reconstructed states are propagated through the model
equations \eqref{eq:edyn} and \eqref{eq:fund}, with $\dot c_i$
additionally requiring $\dot\omega_i^{\mathrm{nom}}$, computable
from $\dot h_i=\omega_i\hpi$ and
$\dot z_{ij}=u_ih_i-u_jh_j$ using the neighbour states
$(u_j,\varphi_j)$ communication. Two consequences
follow. First, \eqref{eq:a_ctrl} never differentiates through the
discontinuous injection term of \eqref{eq:sta2}, so
$a_i^{\mathrm{nom}}$ is well defined at every $t$. Second, for
$t\geq T^*$ and applied inputs equal to the nominal ones, the
substituted rates in \eqref{eq:model_subst} coincide with the true
rates, so \eqref{eq:a_ctrl} is exactly the
smooth, full-information acceleration law achieving
$\dot{\tilde u}_i=-\kappa_3\tilde u_i$; on $[0,T^*)$ it is a bounded,
implementable surrogate, consistent with the two-phase argument of
Remark~\ref{rem:no_exact_claim}.

\begin{assumption}[Gain Condition for Well-Posedness]
\label{ass:kappa1_bound}
Let $\epsilon_d$ and $\epsilon_{\tilde x}$ denote bounds on
$\abs{e_{d,i}(t)}$ and $\abs{\hat{\tilde e}_{x,i}(t)}$ valid for all
$t\ge0$. Such bounds exist and are computable from
\ref{fact:speed}--\ref{fact:range} and Theorem~\ref{thm:sta}; they
are furnished by Lemma~\ref{lem:bounded_transient}, whose proof
invokes only the safety guarantee and is therefore independent of
the value of $\kappa_1$ selected here, so no circular dependence
arises. The gain $\kappa_1>0$ is chosen small enough that
\begin{equation}
  \kappa_1\,(r_{\max}+\epsilon_d)
  \Bigl(\kappa_1\epsilon_d+\frac{\epsilon_{\tilde x}}{r_{\min}}\Bigr)
  \;\le\;\tfrac12,
  \label{eq:kappa1_cond}
\end{equation}
which by \eqref{eq:dF_du} guarantees
$\abs{\partial\hat F/\partial u_i}\leq\tfrac12$, and hence that
\eqref{eq:a_ctrl} is well posed, throughout the operating region.
\end{assumption}

Since the left side of \eqref{eq:kappa1_cond} is continuous and
vanishing as $\kappa_1\to0^+$, a valid $\kappa_1$ always exists;
the condition is a semi-global gain restriction, not a feasibility
assumption.

\section{Safety and Stability Analysis}
\label{sec:stability}

This section completes the design of Fig.~\ref{fig:block_diagram}
and proves its guarantees in a fixed order, chosen so that no step
relies on one that follows it. The HOCBF-QP filter is first
constructed from the reconstructed quantities, with an explicit
tightening that absorbs the reconstruction error. Safety, (P2) and
(P3), is then proved using the filter alone and never the nominal
law; this is what makes \ref{fact:speed}--\ref{fact:range}
available to the control design. Safety in turn bounds the cascade
errors during the estimation transient $[0,T^*)$. Exact exponential
convergence of the cascade is then proved for $t\ge T^*$, and the
main result assembles these pieces.

\subsection{Higher-Order CBF-QP Safety Filter}
\label{ssec:hocbf_qp}

Let us define the barrier candidates
\begin{equation}
  \rho_i\triangleq r_i^2-r_{\min}^2,\quad
  \bar\rho_i\triangleq r_{\max}^2-r_i^2,\quad
  \rho_{ij}\triangleq\norm{z_{ij}}^2-d_{\mathrm{safe}}^2,
  \label{eq:barriers}
\end{equation}
encoding obstacle avoidance, the geofence of \ref{fact:range}, and
inter-agent collision avoidance, respectively. Using
\eqref{eq:e_geometric} and \eqref{eq:fund},
\begin{equation}
  \dot\rho_i=2r_i\dot r_i=2u_i\bar e_{x,i},\qquad
  \dot{\bar\rho}_i=-2u_i\bar e_{x,i},
  \label{eq:rho_dot}
\end{equation}
independent of $(a_i,\omega_i)$: both candidates have relative
degree two, so Definition~\ref{defn:cbf} does not apply directly.
Differentiating once more along
\eqref{eq:unicycle}--\eqref{eq:edyn},
\begin{equation}
  \ddot\rho_i=2u_i^2+2\bar e_{x,i}\,a_i
  +2u_i(\bar e_{y,i}-d_i)\,\omega_i
  =-\ddot{\bar\rho}_i,
  \label{eq:rho_i_ddot}
\end{equation}
affine in $(a_i,\omega_i)$: condition \ref{cond:affine} holds for
both, with the same gain vector up to sign. Throughout this
subsection, the linear class-$\mathcal K_\infty$ choices
$\beta_1(s)=\gamma_0s$, $\beta_2(s)=\gamma_1s$ with
$\gamma_0,\gamma_1>0$ are used, so that
$\psi_{1,i}(x)\triangleq\dot\rho_i+\gamma_0\rho_i$ and the HOCBF
condition of \eqref{eq:hocbf_r2} reads
$\ddot\rho_i+(\gamma_0+\gamma_1)\dot\rho_i+\gamma_0\gamma_1\rho_i
\ge0$, and analogously for $\bar\rho_i$, with
$\bar\psi_{1,i}(x)\triangleq\dot{\bar\rho}_i+\gamma_0\bar\rho_i$
(same gains, for simplicity), and $\rho_{ij}$ (gains
$\gamma_{0,ij},\gamma_{1,ij}>0$).

\begin{proposition}[Forward Invariance of the Speed Box]
\label{prop:speed_cbf}
Let $\rho_{u,i}^-\triangleq u_i-u_{\min}$ and $\rho_{u,i}^+
\triangleq u_{\max}-u_i$, both of relative degree $1$ in $a_i$
since $\dot u_i=a_i$. If $u_i(0)\in[u_{\min},u_{\max}]$
(Assumption~\ref{ass:init}) and the applied acceleration satisfies,
for some $\gamma_u>0$ and all $t\ge0$,
\begin{equation}
  -\gamma_u\rho_{u,i}^-\leq a_i\leq\gamma_u\rho_{u,i}^+,
  \tag{SPD-$i$}
  \label{eq:speed_cbf}
\end{equation}
then, by Corollary~\ref{thm:cbf} applied to each of
$\rho_{u,i}^-,\rho_{u,i}^+$, $u_i(t)\in[u_{\min},u_{\max}]$ for all
$t\geq0$, establishing \ref{fact:speed} unconditionally, since
\eqref{eq:speed_cbf} involves no reconstructed quantity.
\end{proposition}

Under the range-only sensing model, the true $\bar e_{x,i}$,
$\bar e_{y,i}$, and $\dot\rho_i$ in
\eqref{eq:rho_dot}--\eqref{eq:rho_i_ddot} are unavailable; the
filter is implemented with their reconstructions. The \emph{implemented} obstacle
constraint is
\begin{align}
\nonumber 2\hat{\bar e}_{x,i}\,a_i+2u_i(\hat{\bar e}_{y,i}-d_i)\,
  \omega_i\geq&-2u_i^2-2(\gamma_0+\gamma_1)\,u_i\hat{\bar e}_{x,i}\\
  &-\gamma_0\gamma_1\rho_i+\sigma_i(t)\,\bar\Delta_\rho,
  \tag{OBS-$i$}
  \label{eq:hocbf_obs}
\end{align}
and the implemented geofence constraint is its sign-flipped
counterpart
\begin{align}
\nonumber -2\hat{\bar e}_{x,i}\,a_i-2u_i(\hat{\bar e}_{y,i}-d_i)\,
  \omega_i\geq&\;2u_i^2+2(\gamma_0+\gamma_1)\,u_i\hat{\bar e}_{x,i}\\
  &-\gamma_0\gamma_1\bar\rho_i+\sigma_i(t)\,\bar\Delta_\rho,
  \tag{GEO-$i$}
  \label{eq:hocbf_geo}
\end{align}
where $\sigma_i(t)\in\{0,1\}$ is any causal switching signal with
$\sigma_i(t)=1$ for all $t<T_i^*$, realizable either as
$\mathbf1[t<\bar T]$ with the computable bound $\bar T$ of
Remark~\ref{rem:Tstar}, or by the online sliding-detection mechanism
described there; keeping the tightening on longer than $T_i^*$ is
merely conservative, and $\bar\Delta_\rho$ is the robustness margin
constructed in Lemma~\ref{lem:tightening} below.

\begin{lemma}[Tightening Margin]
\label{lem:tightening}
Let $\bar a\triangleq\max(\abs{a_{\min}},a_{\max})$, and let
$\epsilon_x$, $\epsilon_y$ be the transient reconstruction-error
bounds. Set
\begin{equation}
  \bar\Delta_\rho\triangleq
  2\bar a\,\epsilon_x
  +2u_{\max}\omega_{\max}\,\epsilon_y
  +2(\gamma_0+\gamma_1)\,u_{\max}\epsilon_x.
  \label{eq:Delta_def}
\end{equation}
Then, for any $(a_i,\omega_i)\in[a_{\min},a_{\max}]\times
[-\omega_{\max},\omega_{\max}]$ and any $t\ge0$,
satisfaction of \eqref{eq:hocbf_obs} implies satisfaction of the
true HOCBF condition
$\ddot\rho_i+(\gamma_0+\gamma_1)\dot\rho_i+\gamma_0\gamma_1\rho_i
\ge0$, and likewise \eqref{eq:hocbf_geo} implies the true condition
for $\bar\rho_i$.
\end{lemma}
\begin{proof}
Subtracting the implemented left-hand side of \eqref{eq:hocbf_obs}
(moved to one side) from the true condition, the difference is
\begin{equation*}
  -2\tilde e_{x,i}^{\mathrm o}\,a_i
  -2u_i\tilde e_{y,i}^{\mathrm o}\,\omega_i
  -2(\gamma_0+\gamma_1)\,u_i\tilde e_{x,i}^{\mathrm o},
\end{equation*}
using $\bar e_{x,i}=\hat{\bar e}_{x,i}-\tilde e_{x,i}^{\mathrm o}$,
$\bar e_{y,i}=\hat{\bar e}_{y,i}-\tilde e_{y,i}^{\mathrm o}$, and
$\dot\rho_i=2u_i\bar e_{x,i}$. For $t\ge T_i^*$ this difference is
zero (Theorem~\ref{thm:sta}); for $t<T_i^*$ its magnitude is at most
\eqref{eq:Delta_def} by \ref{fact:speed}, the input box, and the
transient bounds $\abs{\tilde e_{x,i}^{\mathrm o}}\le\epsilon_x$,
$\abs{\tilde e_{y,i}^{\mathrm o}}\le\epsilon_y$ of
\eqref{eq:ex_hat} and \eqref{eq:ey_bound}, which is exactly
what the added margin $\sigma_i(t)\bar\Delta_\rho$ absorbs. The
geofence case is identical with all coefficient signs flipped, so
the same bound applies.
\end{proof}

Note that the estimation error enters \eqref{eq:hocbf_obs} in two
places, through $\hat{\dot\rho}_i=2u_i\hat{\bar e}_{x,i}$
\emph{and} through the input coefficients themselves, and
\eqref{eq:Delta_def} accounts for both; the input-coefficient
contribution is bounded only because the QP below confines
$(a_i,\omega_i)$ to the compact input box.

\begin{proposition}[Non-Singularity of the Implemented Gain Vector]
\label{prop:hocbf_nonsingular}
The gain vector
$g_i\triangleq[\,2\hat{\bar e}_{x,i},\;
2u_i(\hat{\bar e}_{y,i}-d_i)\,]$ acting on $(a_i,\omega_i)$ in
\eqref{eq:hocbf_obs}--\eqref{eq:hocbf_geo} satisfies
$\norm{g_i}^2\ge4\,r_{\min}^2\min(1,u_{\min}^2)>0$ for all $t\ge0$.
\end{proposition}
\begin{proof}
By construction \eqref{eq:ey_hat},
$(\hat{\bar e}_{y,i}-d_i)^2=\max(r_i^2-\hat{\bar e}_{x,i}^2,0)$, so
$\hat{\bar e}_{x,i}^2+(\hat{\bar e}_{y,i}-d_i)^2=
\max(r_i^2,\hat{\bar e}_{x,i}^2)\ge r_i^2\ge r_{\min}^2$. Hence
$\norm{g_i}^2=4[\hat{\bar e}_{x,i}^2+u_i^2(\hat{\bar e}_{y,i}
-d_i)^2]\ge4\min(1,u_i^2)\,r_{\min}^2$, and $u_i\ge u_{\min}$ by
Proposition~\ref{prop:speed_cbf}. Condition \ref{cond:nonvanish}
thus holds unconditionally, at every state the reconstruction can
produce.
\end{proof}

For the inter-agent barrier, differentiating $\rho_{ij}$ twice
using $\dot p_i=u_ih_i$ and $\ddot p_i=a_ih_i+u_i\omega_i\hpi$ from
\eqref{eq:heading_vectors} gives
\begin{align}
 \nonumber \ddot\rho_{ij}&=2\norm{\dot z_{ij}}^2
  +2z_{ij}^\top h_i\,a_i+2u_iz_{ij}^\top\hpi\,\omega_i\\
  &\quad-2z_{ij}^\top(a_jh_j+u_j\omega_jh_{\perp,j}),
  \label{eq:hocbf_int_raw}
\end{align}
with $\dot z_{ij}=u_ih_i-u_jh_j$. All quantities here are measured
or communicated: for $j\in\mathcal A_i(t)$, robot $i$ receives
$(u_j,\varphi_j,a_j,\omega_j)$ over the communication link, and robot $j$'s inputs are treated
as given rather than jointly optimized (Remark~\ref{rem:broadcast}).
The gain vector
$[\,2z_{ij}^\top h_i,\;2u_iz_{ij}^\top\hpi\,]$ is nonzero whenever
$z_{ij}\neq0$, since $h_i,\hpi$ form an orthonormal basis of
$\mathbb R^2$ and $u_i\ge u_{\min}>0$: conditions
\ref{cond:affine}--\ref{cond:nonvanish} hold for $\rho_{ij}$. With
$\psi_{1,ij}(x)\triangleq\dot\rho_{ij}+\gamma_{0,ij}\rho_{ij}$, the
implemented constraint, with no estimation-error tightening needed
since no reconstructed quantity appears, is
\begin{align}
\nonumber 2z_{ij}^\top h_i\,a_i{+}2u_iz_{ij}^\top\hpi\,\omega_i
  \geq&{-}2\norm{\dot z_{ij}}^2
  +2z_{ij}^\top(a_jh_j{+}u_j\omega_jh_{\perp,j})\\
 &-\gamma_{0,ij}\dot\rho_{ij}-\gamma_{1,ij}\psi_{1,ij}.
  \tag{INT-$ij$}
  \label{eq:hocbf_int}
\end{align}
Constraint \eqref{eq:hocbf_int} is enforced over the time-varying
set $\mathcal A_i(t)$ of \eqref{eq:active_set}, i.e., for the
always-measured ring neighbours and for any other robot currently
within sensing range.

\begin{remark}[Causal Use of Communicated Inputs]
\label{rem:broadcast}
Robots $i$ and $j$ solve their QPs simultaneously, each needing the
other's current inputs in \eqref{eq:hocbf_int}. In the sampled-data
implementation of Section~\ref{sec:simulation}, the inputs are
held constant over each sampling interval, and the communicated
$(a_j,\omega_j)$ are those robot $j$ applied over the preceding
interval; the pair $(i,j)$ thus enforces \eqref{eq:hocbf_int} from
both sides with a one-sample delay in the other robot's input,
which is causal and requires no joint optimization. The
continuous-time analysis below treats $(a_j,\omega_j)$ as the
inputs robot $j$ actually applies.
\end{remark}

Because non-neighbour constraints switch on at
$\norm{z_{ij}}=R_s$, Corollary~\ref{thm:hocbf} must hold from the
activation instant, which requires $\psi_{1,ij}\ge0$ there. This is
guaranteed by a checkable gain condition:

\begin{assumption}[Activation and Safe Initialization]
\label{ass:safe_init}
(a) The inter-agent gains satisfy
\begin{equation}
  \gamma_{0,ij}\;\ge\;\frac{4R_s\,u_{\max}}{R_s^2
  -d_{\mathrm{safe}}^2},\qquad \forall\,i\ne j.
  \label{eq:activation_gain}
\end{equation}
(b) At $t=0$: $\gamma_0\rho_i(0)\ge2r_i(0)u_{\max}$ and
$\gamma_0\bar\rho_i(0)\ge2r_i(0)u_{\max}$ for all $i$, and
$\gamma_{0,ij}\rho_{ij}(0)\ge4\norm{z_{ij}(0)}u_{\max}$ for all
pairs with $\norm{z_{ij}(0)}\le R_s$.
\end{assumption}

Part (a) guarantees $\psi_{1,ij}\ge0$ whenever a pair enters
sensing range: at $\norm{z_{ij}}=R_s$,
$\rho_{ij}=R_s^2-d_{\mathrm{safe}}^2$ while $\dot\rho_{ij}=
2z_{ij}^\top\dot z_{ij}\ge-4R_su_{\max}$ by \ref{fact:speed}, so
$\psi_{1,ij}\ge\gamma_{0,ij}(R_s^2-d_{\mathrm{safe}}^2)
-4R_su_{\max}\ge0$. Part (b) is the analogous condition at $t=0$
for the constraints active from the start; each inequality is a
sufficient, measurement-checkable version of
$\psi_1(0)\ge0$, obtained by bounding the unknown initial rates
through $\abs{\dot r_i(0)}\le u_{\max}$ and
$\abs{\dot\rho_{ij}(0)}\le4\norm{z_{ij}(0)}u_{\max}$.

The safety filter solves, for each robot $i$ at each instant $t$,
\begin{equation}
\begin{aligned}
  (a_i^\star,\omega_i^\star)=\arg\min_{(a_i,\omega_i)}\;&
  \tfrac12(a_i-a_i^{\mathrm{nom}})^2+
  \tfrac12(\omega_i-\omega_i^{\mathrm{nom}})^2\\
  \text{s.t.}\;\;&\text{\eqref{eq:hocbf_obs},
  \eqref{eq:hocbf_geo}, \eqref{eq:speed_cbf}},\\
  &\text{\eqref{eq:hocbf_int} for all } j\in\mathcal A_i(t),\\
  &a_i\in[a_{\min},a_{\max}],\ \
  \omega_i\in[-\omega_{\max},\omega_{\max}].
\end{aligned}
\label{eq:QP}
\end{equation}
The objective is strictly convex, so \eqref{eq:QP} has a unique
solution whenever feasible, and it returns
$(a_i^{\mathrm{nom}},\omega_i^{\mathrm{nom}})$ exactly whenever the
nominal pair itself satisfies every constraint, the
minimal-invasiveness property central to the design.

\begin{assumption}[Actuator Authority]
\label{ass:actuator}
$a_{\min},a_{\max}$, and $\omega_{\max}$ are such that, at every
$t\geq0$, the constraints of \eqref{eq:QP}, jointly, over the
set $\mathcal A_i(t)$, and including the worst case
$\sigma_i(t)=1$, admit a common feasible $(a_i,\omega_i)$ in the
input box.
\end{assumption}

\subsection{Unconditional Safety Through the Estimation Transient}
\label{ssec:safety_thm}

\begin{theorem}[Unconditional Safety]
\label{thm:safety}
Under Assumptions~\ref{ass:init}, \ref{ass:jerk},
\ref{ass:safe_init}, and \ref{ass:actuator}, if
$(a_i,\omega_i)=(a_i^\star,\omega_i^\star)$ solves \eqref{eq:QP} for
every $i$ and every $t\geq0$, then for all $t\ge0$:
$u_i(t)\in[u_{\min},u_{\max}]$ and
$r_{\min}\le r_i(t)\le r_{\max}$ for all $i$, establishing
\ref{fact:speed}--\ref{fact:range}, and
$\norm{p_i(t)-p_j(t)}\ge d_{\mathrm{safe}}$ for all $i\ne j$; i.e.,
(P2) and (P3) of Problem~\ref{prob:main} hold, regardless of the
differentiator's transient error on $[0,T^*)$.
\end{theorem}
\begin{proof}
The speed box is Proposition~\ref{prop:speed_cbf}. For the obstacle
and geofence barriers: by Lemma~\ref{lem:tightening}, the enforced
constraints \eqref{eq:hocbf_obs}--\eqref{eq:hocbf_geo} imply the
true HOCBF conditions on $\rho_i$ and $\bar\rho_i$ at every
$t\ge0$, exactly for $t\ge T_i^*$ and through the tightening margin
for $t<T_i^*$; Assumption~\ref{ass:safe_init}(b) gives
$\psi_{1,i}(0)\ge0$ and $\bar\psi_{1,i}(0)\ge0$, and
Proposition~\ref{prop:hocbf_nonsingular} with
Assumption~\ref{ass:actuator} gives solvability, so
Corollary~\ref{thm:hocbf} yields $\rho_i(t)\ge0$ and
$\bar\rho_i(t)\ge0$ for all $t$. For inter-agent safety, fix
$i\ne j$. While $j\in\mathcal A_i(t)$, constraint
\eqref{eq:hocbf_int} enforces the true HOCBF condition on
$\rho_{ij}$ directly, with
$\psi_{1,ij}\ge0$ holding at $t=0$ by
Assumption~\ref{ass:safe_init}(b) or at each activation instant by
Assumption~\ref{ass:safe_init}(a), so Corollary~\ref{thm:hocbf} gives
$\rho_{ij}\ge0$ on every interval of activity. While
$j\notin\mathcal A_i(t)$, $\norm{z_{ij}}>R_s>d_{\mathrm{safe}}$ by
definition of $\mathcal A_i(t)$, so the constraint is satisfied
trivially, and any approach toward collision must first cross
$\norm{z_{ij}}=R_s$, re-entering the constrained regime.
Concatenating the two regimes gives $\rho_{ij}(t)\ge0$ for all
$t\ge0$.
\end{proof}

\subsection{Boundedness on the Estimation Transient}
\label{ssec:bounded}

With safety established, the cascade errors are automatically
confined:

\begin{lemma}[Uniform Boundedness of the Cascade Errors]
\label{lem:bounded_transient}
Under the hypotheses of Theorem~\ref{thm:safety}, the cascade errors
$e_{d,i}(t)$, $\hat{\tilde e}_{x,i}(t)$, $\tilde u_i(t)$ are bounded
for all $t\ge0$, in particular on $[0,T^*)$, by computable
constants $\epsilon_d$, $\epsilon_{\tilde x}$, $\epsilon_{\tilde u}$
depending only on quantities fixed in
Sections~\ref{sec:problem}--\ref{sec:control}.
\end{lemma}
\begin{proof}
By Theorem~\ref{thm:safety}, $r_{\min}\le r_i\le r_{\max}$ and
$u_i\in[u_{\min},u_{\max}]$ for all $t\ge0$; hence
$\abs{e_{d,i}}\le\epsilon_d\triangleq\max_i\max(r_{\max}-d_i,\,
d_i-r_{\min})$ and $\norm{z_{ij}}\le r_i+r_j\le2r_{\max}$; the
geofence makes the neighbour-displacement bound a consequence, not
an assumption. The differentiator's strict Lyapunov function
provides an invariant sublevel set for
$(\eta_{i,1},\eta_{i,2})$~\cite{moreno2012strict}, so
$\abs{\zeta_{i,2}}\le\abs{\dot r_i}+\abs{\eta_{i,2}}\le
u_{\max}+\epsilon_2$, giving
$\abs{\hat{\bar e}_{x,i}}=\abs{r_i\zeta_{i,2}/u_i}\le
r_{\max}(u_{\max}+\epsilon_2)/u_{\min}$ and hence a bound
$\epsilon_{\tilde x}$ on $\hat{\tilde e}_{x,i}=\hat{\bar e}_{x,i}+
\kappa_1r_ie_{d,i}$. From \eqref{eq:ey_hat},
$\abs{\hat{\bar e}_{y,i}-d_i}\le r_i\le r_{\max}$; from
\eqref{eq:phase_law} and the bound on $\norm{z_{ij}}$,
$\omega_i^{\mathrm{nom}}$ is bounded, hence so is $c_i$ in
\eqref{eq:ci_def}; and $u_i^\star=\hat F(\cdot)$ in \eqref{eq:F_def}
is a finite sum of products of bounded quantities, bounding
$\tilde u_i=u_i-u_i^\star$ by some $\epsilon_{\tilde u}$. No step
differentiated through the sliding term.
\end{proof}

Lemma~\ref{lem:bounded_transient} supplies exactly the constants
$\epsilon_d,\epsilon_{\tilde x}$ against which
Assumption~\ref{ass:kappa1_bound} calibrates $\kappa_1$.

\subsection{Exact Cascade Convergence for $t\geq T^*$}
\label{ssec:cascade_thm}

Let $\mathcal E_\Phi\triangleq\{\theta:\nabla\Phi(\theta)=0\}$
denote the critical set of the formation potential \eqref{eq:Vf}.

\begin{theorem}[Cascade Convergence]
\label{thm:cascade}
Let the hypotheses of Theorem~\ref{thm:safety} and
Assumption~\ref{ass:kappa1_bound} hold, and suppose
$(a_i,\omega_i)=(a_i^{\mathrm{nom}},\omega_i^{\mathrm{nom}})$ for
every $i$ and every $t\geq T^*$. If the gains satisfy
\begin{equation}
  \kappa_2\;>\;\frac12+\frac{u_{\max}}{2\kappa_1r_{\min}^2},
  \qquad
  \kappa_3\;>\;\frac12,
  \label{eq:gain_cond}
\end{equation}
then, with $W\triangleq\tfrac12\sum_i\bigl(e_{d,i}^2+
\tilde e_{x,i}^2+\tilde u_i^2\bigr)$ evaluated for $t\geq T^*$
(where $\hat{\tilde e}_{x,i}=\tilde e_{x,i}$ exactly),
\begin{equation}
  \dot W\leq-2\lambda W,\qquad
  \lambda\triangleq\min\Bigl(\tfrac{\kappa_1u_{\min}}2,\;
  \kappa_2-\tfrac12-\tfrac{u_{\max}}{2\kappa_1r_{\min}^2},\;
  \kappa_3-\tfrac12\Bigr),
  \label{eq:W_cascade}
\end{equation}
so $(e_{d,i},\tilde e_{x,i},\tilde u_i)\to(0,0,0)$ exponentially
with rate at least $\lambda$, from the bounded state
Lemma~\ref{lem:bounded_transient} guarantees at $t=T^*$; hence
$r_i(t)\to d_i$ exactly as $t\to\infty$. On the limit
manifold $\alpha_i\equiv\pi/2$, $r_i\equiv d_i$ reached by this
convergence, and per the scope of Remark~\ref{rem:phase_scope},
$\theta(t)$ converges locally exponentially to $\mathcal E_\Phi$;
convergence over the full transient, rather than on this reduced
manifold, remains open.
\end{theorem}
\begin{proof}
For $t\geq T^*$, $\tilde e_{x,i}^{\mathrm o}=
\tilde e_{y,i}^{\mathrm o}=0$, so \eqref{eq:ed_stage1} holds
with no reconstruction-error term, the substituted rates
\eqref{eq:model_subst} coincide with the true ones, and by
construction of \eqref{eq:u_star} and \eqref{eq:a_ctrl},
\begin{equation*}
  \dot e_{d,i}=-\kappa_1u_ie_{d,i}+\frac{u_i}{r_i}\tilde e_{x,i},
  \quad
  \dot{\tilde e}_{x,i}=-\kappa_2\tilde e_{x,i}+\tilde u_i,
  \quad
  \dot{\tilde u}_i=-\kappa_3\tilde u_i,
\end{equation*}
hold exactly. Differentiating $W$ term-wise: for the radial channel,
\begin{align*}
  -\kappa_1u_ie_{d,i}^2+\frac{u_i}{r_i}e_{d,i}\tilde e_{x,i}
  &\le u_i\Bigl[-\frac{\kappa_1}2e_{d,i}^2
  +\frac{\tilde e_{x,i}^2}{2\kappa_1r_{\min}^2}\Bigr]\\
  &\le-\frac{\kappa_1u_{\min}}2e_{d,i}^2
  +\frac{u_{\max}}{2\kappa_1r_{\min}^2}\tilde e_{x,i}^2,
\end{align*}
by Young's inequality with $r_i\ge r_{\min}$ (Theorem
\ref{thm:safety}) and then $u_i\in[u_{\min},u_{\max}]$ applied to
the negative and positive parts respectively; and
$\tilde e_{x,i}\tilde u_i\le\tfrac12\tilde e_{x,i}^2+
\tfrac12\tilde u_i^2$. Collecting terms gives
$\dot W\le-\frac{\kappa_1u_{\min}}2\sum_ie_{d,i}^2
-(\kappa_2-\frac12-\frac{u_{\max}}{2\kappa_1r_{\min}^2})
\sum_i\tilde e_{x,i}^2-(\kappa_3-\frac12)\sum_i\tilde u_i^2$, which
is \eqref{eq:W_cascade} under \eqref{eq:gain_cond}. The
$\theta$-statement is not a consequence of $W$: per
Remark~\ref{rem:phase_scope}, the descent property of
\eqref{eq:phase_law} holds on the manifold
$\alpha_i\equiv\pi/2$, $r_i\equiv d_i$, and there the local claim
follows from the Hessian $\nabla^2\Phi(\theta^\star)$, positive
definite transverse to uniform rotation, via the standard
linearization argument for gradient-type flows at a strict local
minimum.
\end{proof}

\subsection{Main Result}
\label{ssec:main_result}

\begin{theorem}[Safety with Asymptotically Exact Convergence]
\label{thm:full_result}
Under Assumptions~\ref{ass:init}, \ref{ass:jerk},
\ref{ass:kappa1_bound}, \ref{ass:safe_init}, \ref{ass:actuator},
and the gain condition \eqref{eq:gain_cond}, the closed loop under
\eqref{eq:QP}:
\begin{enumerate}[label=\textnormal{(\roman*)}]
\item satisfies, for all $t\geq0$ and all $i\ne j$:
  $r_{\min}\le r_i(t)\le r_{\max}$,
  $\norm{p_i(t)-p_j(t)}\ge d_{\mathrm{safe}}$, and
  $u_i(t)\in[u_{\min},u_{\max}]$ by Theorem~\ref{thm:safety},
  proved independently of the cascade and hence holding even while
  it is still transient;
\item recovers the nominal law on every time interval
  $\mathcal I\subseteq[T^*,\infty)$ on which no constraint of
  \eqref{eq:QP} is active:
  $(a_i^\star,\omega_i^\star)=(a_i^{\mathrm{nom}},
  \omega_i^{\mathrm{nom}})$ exactly on $\mathcal I$, so the
  exponential decay \eqref{eq:W_cascade} of
  Theorem~\ref{thm:cascade} holds on $\mathcal I$, started from
  the bounded state of Lemma~\ref{lem:bounded_transient};
\item consequently, if there exists $T_f\ge T^*$ after which no
  constraint of \eqref{eq:QP} is active, achieves $r_i(t)\to d_i$
  exactly as $t\to\infty$ and, on the manifold this reaches, local
  convergence of $\theta(t)$ to $\mathcal E_\Phi$ per
  Theorem~\ref{thm:cascade}, with any radial deviation confined to
  the finite window $[0,T_f)$ during which the differentiator has
  not converged or a safety constraint is transiently active, and
  never a safety violation, by (i). Formation convergence over the
  full transient is verified numerically in
  Section~\ref{sec:simulation} and remains analytically open, per
  Remark~\ref{rem:phase_scope}.
\end{enumerate}
\end{theorem}
\begin{proof}
Item (i) is Theorem~\ref{thm:safety}. For (ii), fix $t\in\mathcal I$;
the unconstrained minimizer of the strictly convex objective
coincides with its target whenever the target is feasible, so
$(a_i^\star,\omega_i^\star)=(a_i^{\mathrm{nom}},
\omega_i^{\mathrm{nom}})$, which is the hypothesis of
Theorem~\ref{thm:cascade} on $\mathcal I$;
Lemma~\ref{lem:bounded_transient} supplies the bounded state at the
start of $\mathcal I$, holding independently of which constraints
were active before. Item (iii) is (ii) with
$\mathcal I=[T_f,\infty)$ combined with (i).
\end{proof}

\section{Simulation Results}
\label{sec:simulation}

\begin{figure*}[t]
\centering
\begin{subfigure}{0.3\linewidth}
  \centering
  \includegraphics[width=0.925\linewidth]{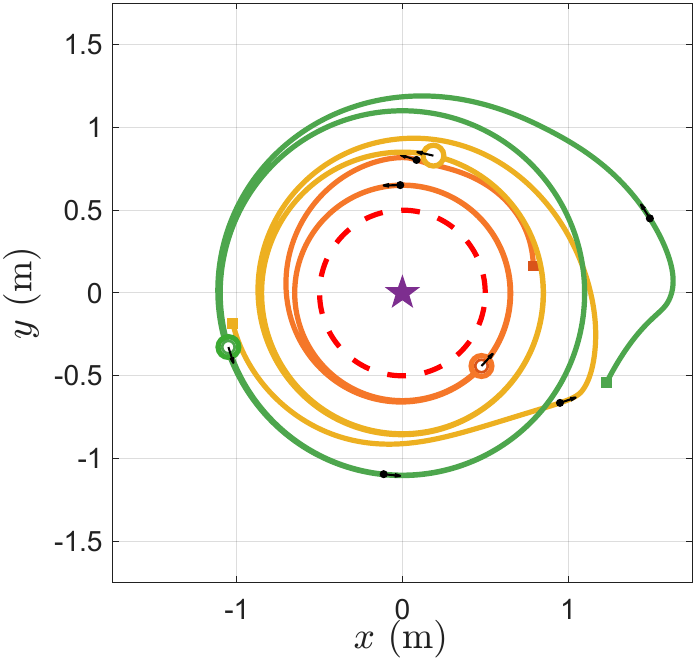}
  \caption{Trajectories}
\end{subfigure}
\begin{subfigure}{0.3\linewidth}
  \centering 
  \includegraphics[width=\linewidth]{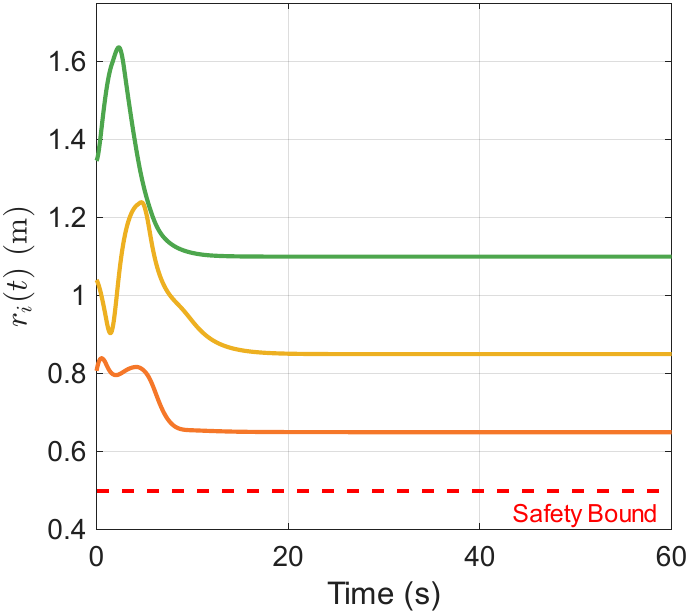}
  \caption{Range evolution}
\end{subfigure}
\begin{subfigure}{0.3\linewidth}
  \centering
  \includegraphics[width=\linewidth]{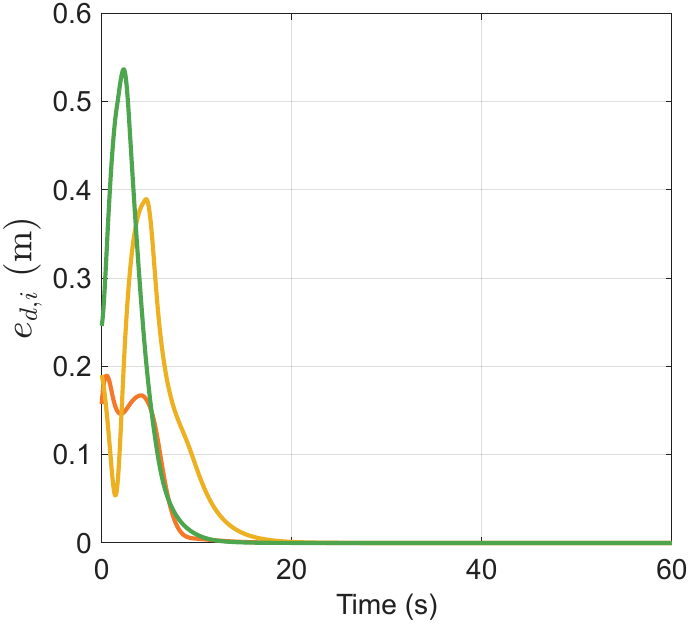}
  \caption{Radial error}
\end{subfigure}
\caption{Trajectories and time evolution of range and radial error
under the proposed HOCBF-QP controller~\eqref{eq:QP} for $N=3$
robots. All robots converge to their heterogeneous orbits while
remaining outside the safety boundary, and $e_{d,i}(t)$
decays to zero for every robot.}
\vspace{-0.25cm}
\label{fig:Range_only_syn2}
\end{figure*}

\begin{figure*}[t]
\centering
\begin{subfigure}{0.3\linewidth}
  \centering
  \includegraphics[width=0.925\linewidth]{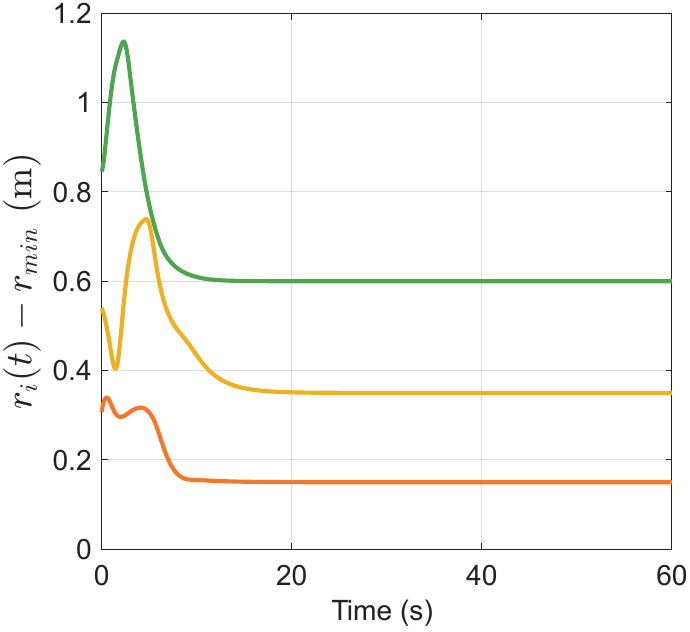}
  \caption{Distance from obstacle}
\end{subfigure}
\begin{subfigure}{0.3\linewidth}
  \centering 
  \includegraphics[width=\linewidth]{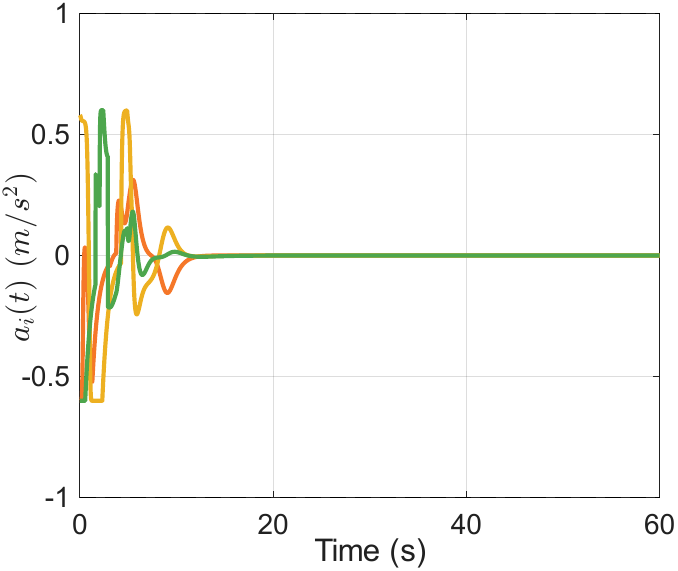}
  \caption{Acceleration control}
\end{subfigure}
\begin{subfigure}{0.3\linewidth}
  \centering 
  \includegraphics[width=\linewidth]{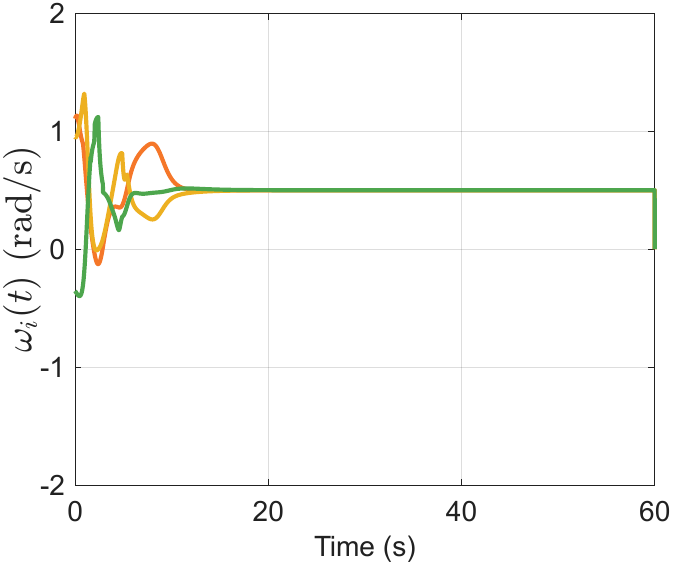}
  \caption{Turn-rate control}
\end{subfigure}
\caption{Obstacle-safety margin and control inputs under the
proposed HOCBF-QP controller~\eqref{eq:QP} for $N=3$ robots. Each
robot's distance from the obstacle boundary stays strictly positive
throughout, confirming obstacle avoidance, while the acceleration
and turn-rate commands settle from an initial transient to the
steady values sustaining the converged orbits.}
\vspace{-0.25cm}
\label{fig:Range_only_syn}
\end{figure*}

\subsection{Setup}
\label{ssec:sim_setup}

The proposed HOCBF-QP controller~\eqref{eq:QP} is validated on
$N=3$ unicycle robots assigned heterogeneous orbital radii
$d=(0.65,0.85,1.10)$~m around an obstacle-target of radius
$r_{\min}=0.50$~m at the origin, with inter-agent margin
$d_{\mathrm{safe}}=0.25$~m, geofence radius $r_{\max}=1.70$~m, and
sensing and communication radius $R_s=6d_{\mathrm{safe}}=1.50$~m.
The team size $N=3$ is that of the baseline
of~\cite{zheng2015distributed} used in the comparison; the controller itself depends on
$N$ only through the chord lengths~\eqref{eq:chord}, and each
robot uses the data of its two ring neighbours alone.
Robots start off their orbits with randomized radii, phases, and
headings (fixed seed) and speeds $u_i(0)$ at the midpoint of
$[u_{\min},u_{\max}]=[0.05,1.50]$~m/s; every barrier active at
$t=0$ satisfies $\psi_1(0)\ge0$, the condition that
Assumption~\ref{ass:safe_init}(b) is designed to guarantee. Actuator limits are
$a_i\in[-1.0,1.0]$~m/s$^2$ and $\omega_i\in[-3.0,3.0]$~rad/s.

The nominal law uses gains $\kappa_1=1.0$, $\kappa_2=2.0$,
$\kappa_3=5.0$, $\kappa_w=0.5$, and phase-advance rate
$\omega_0=0.5$~rad/s, with $\dot u_i^\star$ in~\eqref{eq:a_ctrl}
produced by a dynamic-surface filter of time constant
$\tau_f=0.05$~s. The HOCBF-QP filter uses
$\gamma_0=\gamma_1=\gamma_{0,ij}=\gamma_{1,ij}=0.8$, kept small so
the filter shapes the trajectory before a boundary is reached
rather than correcting abruptly at it; the super-twisting gains
follow the rule of~\cite{moreno2012strict} for $L$
in~\eqref{eq:L_def}, with $\sigma_i(t)$ from the online sliding
detection of Remark~\ref{rem:Tstar}. At each step \eqref{eq:QP} is
solved per robot by a projected-gradient dual solver, with the
communicated neighbour inputs $(a_j,\omega_j)$ taken from the
preceding step as in Remark~\ref{rem:broadcast}, and integrated
through the true dynamics~\eqref{eq:unicycle} by fourth-order
Runge--Kutta with $\Delta t=0.01$~s over $60$~s.

The gain conditions
are sufficient, worst-case bounds, and the simulations deliberately
probe beyond them: with the values above, \eqref{eq:gain_cond}
would require $\kappa_2>3.5$ and \eqref{eq:kappa1_cond} a
considerably smaller $\kappa_1$, while the inter-agent gains lie
below the bound~\eqref{eq:activation_gain} of
Assumption~\ref{ass:safe_init}(a). The behaviour the theory
predicts is nonetheless observed: $\psi_{1,ij}\ge0$ held at every
activation instant, the geofence never activated, and the cascade
converged exponentially. This is the usual conservatism of
Lyapunov-based gain bounds, not a weakening of the safety
guarantee, which Theorem~\ref{thm:safety} ties to the QP
constraints rather than to the cascade gains.

\subsection{Convergence and Safety of the Proposed Controller}

Fig.~\ref{fig:Range_only_syn2} reports trajectories, ranges, and
radial error. All three robots settle from their randomized starts
onto concentric, non-intersecting orbits at the assigned radii
$d_i$, and each range $r_i(t)$ converges to its target without ever
approaching the boundary $r_{\min}$. The radial error
$e_{d,i}(t)$ rises to a brief peak, as the initial offset combines
with the time the differentiator of
Section~\ref{ssec:differentiator} takes to lock onto the range
rate, then decays to numerical zero well within the horizon,
consistent with the exponential rate of Theorem~\ref{thm:cascade}.

Fig.~\ref{fig:Range_only_syn} reports the obstacle margin and the
two control channels. The margin $r_i(t)-r_{\min}$ dips during the
initial transient but stays strictly positive throughout, in
agreement with Theorem~\ref{thm:safety}, and does so even while the
differentiator is still converging, confirming that the
tightening~\eqref{eq:hocbf_obs} does its job. The acceleration and
turn-rate commands show a brief transient, sharper on the
acceleration channel while~\eqref{eq:hocbf_obs} is intermittently
binding, then settle to steady values, with the turn-rate near
$\omega_0$ for every robot, consistent with a balanced ring at the
common phase-advance rate.

\begin{figure}[t]
\centering
\begin{subfigure}{0.435\linewidth}
  \centering
  \includegraphics[width=\linewidth]{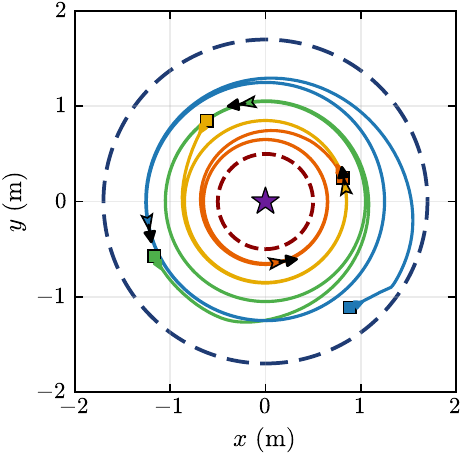}
  \caption{Trajectories}
  \label{fig:N4_traj}
\end{subfigure}\hfill
\begin{subfigure}{0.55\linewidth}
  \centering
  \includegraphics[width=\linewidth]{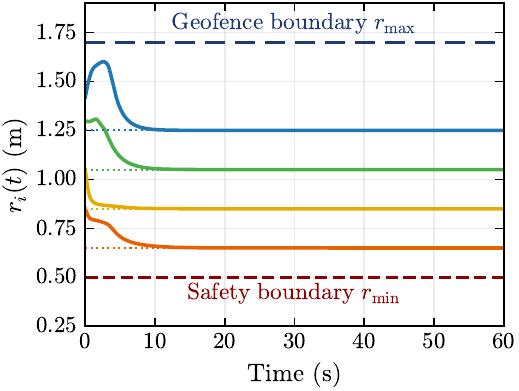}
  \caption{Range evolution}
  \label{fig:N4_range}
\end{subfigure}
\caption{Proposed controller with $N=4$ robots on the sparse ring
graph, where each robot coordinates with only two of its three
teammates.}
\vspace{-0.25cm}
\label{fig:N4}
\end{figure}

\begin{figure}[t]
\centering
\begin{subfigure}{0.35\linewidth}
  \centering
  \includegraphics[width=\linewidth]{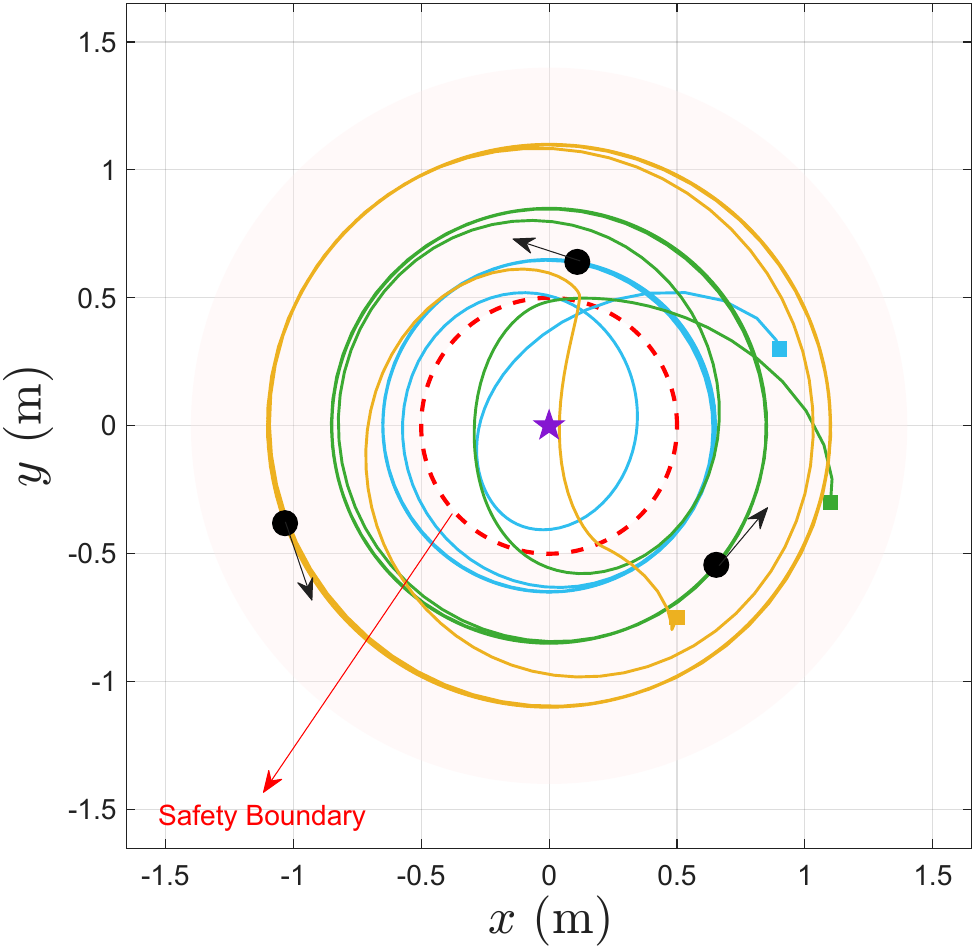}
  \caption{Trajectories}
  \label{fig:traj_comp}
\end{subfigure}\hfill
\begin{subfigure}{0.55\linewidth}
  \centering
  \includegraphics[width=\linewidth]{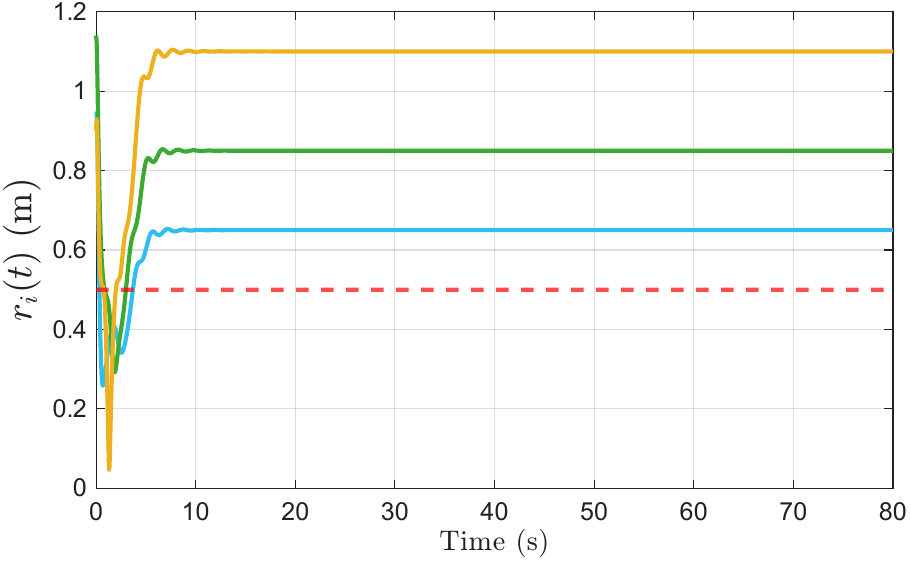}
  \caption{Range to target}
  \label{fig:safety_comp}
\end{subfigure}
\caption{Baseline of \cite{zheng2015distributed} on the $N=3$. The trajectories in
(a) crosses the safety disc (red dashed) and cross one another
near the target, and the ranges in (b) lie below $r_{\min}$ for the
first seconds of the run before settling on the assigned orbits.}
\vspace{-0.25cm}
\label{fig:comparison}
\end{figure}

For $N=3$ the ring graph coincides with the complete graph, so the
scenario above does not exercise the sparsity of the coordination
topology. Fig.~\ref{fig:N4} repeats the experiment with
$N=4$ robots on radii $d=(0.65,0.85,1.05,1.25)$~m, started from
$r_i(0)=(0.85,1.05,1.30,1.42)$~m,
$\theta_i(0)=(0.30,2.20,3.60,-0.90)$~rad, and
$\alpha_i(0)=(1.70,2.00,1.60,1.30)$~rad, with all other parameters
unchanged. Each robot now coordinates with its two ring neighbours
only. All four ranges converge to their assigned radii, the trajectories settle onto non-intersecting
orbits, and both the standoff and the geofence margins stay positive
throughout, confirming that the design operates on a graph strictly
sparser than the complete one.

Comparison with \cite{zheng2015distributed}. To quantify the safety gap of Section~\ref{sec:intro}, the closest
heterogeneous multi-orbit design~\cite{zheng2015distributed} is run
unmodified on the same $N=3$ configuration, with the same target,
radii $d_i$, and margins $r_{\min}$ and $d_{\mathrm{safe}}$, over an
$80$~s horizon: its own velocity-level inputs, virtual-agent
normalization $\delta_i=1/d_i$, parameters
$a_i^{\mathrm{Z}}=1-c\,(d_i/d_{i+1})$ with $c=-0.9$, and gains
$k_v=2.75$, $k_\omega=3.897$, the superscript $\mathrm{Z}$
distinguishing the baseline's ratio parameter from the acceleration
input $a_i$ of this paper. Unlike the proposed controller, it is
granted full state access, absolute positions and velocities, yet
carries no safety mechanism.

Fig.~\ref{fig:comparison} shows the outcome. The baseline does reach
the assigned radii, but only after violating safety in both
directions. In Fig.~\ref{fig:safety_comp}, all three ranges lie
below $r_{\min}=0.50$~m during the first seconds of the run, one
agent passing very close to the target, and they cross the standoff
boundary only as the orbits are approached. In
Fig.~\ref{fig:traj_comp}, the corresponding trajectories cut through
the safety disc and intersect one another near the target, where the
pairwise distances fall well below $d_{\mathrm{safe}}=0.25$~m.
Neither violation is a tuning artefact, the gains being the
baseline's published values; it is a structural consequence of
certifying convergence alone. The proposed controller, by contrast,
reaches the same orbits on the same configuration with both margins
strictly positive at every instant
(Figs.~\ref{fig:Range_only_syn2}--\ref{fig:Range_only_syn}), under
range-only sensing and with no orbit-dependent tuning.

\section{Conclusion}
\label{sec:conc}

This paper presented a safety-critical controller for unicycle
robots circumnavigating an obstacle-target on heterogeneous orbits,
using both acceleration and turn-rate. A single
control law serves every robot regardless of its assigned radius,
because the body-frame radial error does not depend on that radius.
A backstepping law in the acceleration channel drives each radius
exactly to its target, while the turn-rate channel balances the
formation with the ring neighbours. Obstacle, geofence, and
inter-agent safety are enforced by a HOCBF-QP acting on both inputs, which leaves the nominal law
untouched away from the constraints and, through a finite-time
differentiator and an explicit constraint tightening, remains valid
while the range rate is still being estimated. Simulations
confirmed these properties and showed that the baseline
of~\cite{zheng2015distributed}, with full state access, violates
both the target standoff and the inter-agent margin where the
proposed controller does not. Future work will address the
formation analysis over the full transient, time-varying graphs, moving targets, and hardware validation.

\bibliographystyle{IEEEtran}
\bibliography{ref}
    
\end{document}